\documentclass{article}

\usepackage{arxiv}

\usepackage[utf8]{inputenc} % allow utf-8 input
\usepackage[T1]{fontenc}    % use 8-bit T1 fonts
\usepackage{hyperref}       % hyperlinks
\usepackage{url}            % simple URL typesetting
\usepackage{booktabs}       % professional-quality tables
\usepackage{amsfonts}       % blackboard math symbols
\usepackage{nicefrac}       % compact symbols for 1/2, etc.
\usepackage{microtype}      % microtypography
\usepackage{lipsum}         % Can be removed after putting your text content
\usepackage{graphicx}
\usepackage{doi}
\usepackage{natbib}

\usepackage{multirow}
\usepackage{amssymb}
\usepackage{amsmath}
\usepackage{tabularx}
\usepackage[ruled]{algorithm}
\usepackage[noend]{algpseudocode}
\usepackage{subcaption}
\usepackage{cleveref}       % smart cross-referencing

\hypersetup{
	colorlinks=true,
	citecolor=blue,   % color of citation links
	linkcolor=blue,  % color of internal links (sections, equations)
	urlcolor=blue     % color of URLs
}

\expandafter\def\expandafter\normalsize\expandafter{%
  \normalsize  
  \setlength\abovedisplayskip{-10pt}
  \setlength\belowdisplayskip{8pt}
  \setlength\abovedisplayshortskip{-10pt}
  \setlength\belowdisplayshortskip{8pt}
}
\title{Evaluating the stability of model explanations in instance-dependent cost-sensitive credit scoring}

\date{}

\newif\ifuniqueAffiliation
\uniqueAffiliationtrue

\ifuniqueAffiliation % Standard variant of author block
\author{ \href{https://orcid.org/0009-0006-0881-5644}{\includegraphics[scale=0.06]{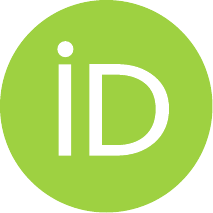}\hspace{1mm}Matteo Ballegeer}\thanks{Corresponding author} \\
	Ghent University\\
	CVAMO Core Lab\\
	\texttt{matteo.ballegeer@ugent.be} \\
	\And
	\href{https://orcid.org/0000-0002-4502-0764}{\includegraphics[scale=0.06]{orcid.pdf}\hspace{1mm}Matthias Bogaert} \\
	Ghent University\\
	CVAMO Core Lab\\
	\texttt{matthias.bogaert@ugent.be} \\
	\And
	\href{https://orcid.org/0000-0001-9901-8507}{\includegraphics[scale=0.06]{orcid.pdf}\hspace{1mm}Dries F. Benoit} \\
	Ghent University\\
	CVAMO Core Lab\\
	\texttt{dries.benoit@ugent.be} \\
}
\else
\usepackage{authblk}

\newbox{\orcid}\sbox{\orcid}{\includegraphics[scale=0.06]{orcid.pdf}}
\author[1,2]{%
	\href{https://orcid.org/0009-0006-0881-5644}{\usebox{\orcid}\hspace{1mm}Matteo Ballegeer\thanks{\texttt{matteo.ballegeer@ugent.be}}}%
}
\author[1,2]{%
	\href{https://orcid.org/0000-0002-4502-0764}{\usebox{\orcid}\hspace{1mm}Matthias Bogaert\thanks{\texttt{matthias.bogaert@ugent.be}}}%
}
\author[1,2]{%
	\href{https://orcid.org/0000-0001-9901-8507}{\usebox{\orcid}\hspace{1mm}Dries F. Benoit\thanks{\texttt{dries.benoit@ugent.be}}}%
}
\affil[1]{Ghent University, Research Group Data Analytics, Faculty of Economics and Business Administration, Tweekerkenstraat 2, Belgium}
\affil[2]{FlandersMake@UGent---Corelab CVAMO, Tweekerkenstraat 2, Belgium}
\fi

\renewcommand{\headeright}{}
\renewcommand{\undertitle}{ACCEPTED MANUSCRIPT VERSION OF AN ARTICLE PUBLISHED IN THE EUROPEAN JOURNAL OF OPERATIONAL
RESEARCH (10.1016/j.ejor.2025.05.039)}
\renewcommand{\shorttitle}{\footnotesize ACCEPTED MANUSCRIPT VERSION, PUBLISHED IN THE EUROPEAN JOURNAL OF OPERATIONAL RESEARCH}

\hypersetup{
pdftitle={Evaluating the stability of model explanations in instance-dependent cost-sensitive credit scoring},
pdfsubject={cs.LG,q-fin.ST},
pdfauthor={Matteo Ballegeer, Matthias Bogaert, Dries F. Benoit},
pdfkeywords={Analytics, Cost-sensitive learning, Credit scoring, Explainable AI, Explanation stability},
}

\begin{document}
\maketitle
\begin{abstract}
Instance-dependent cost-sensitive (IDCS) classifiers offer a promising approach to improving cost-efficiency in credit scoring by tailoring loss functions to instance-specific costs. 
However, the impact of such loss functions on the stability of model explanations remains unexplored in literature, despite increasing regulatory demands for transparency. 
This study addresses this gap by evaluating the stability of Local Interpretable Model-agnostic Explanations (LIME) and SHapley Additive exPlanations (SHAP) when applied to IDCS models. 
Using four publicly available credit scoring datasets, we first assess the discriminatory power and cost-efficiency of IDCS classifiers, introducing a novel metric to enhance cross-dataset comparability.
We then investigate the stability of SHAP and LIME feature importance rankings under varying degrees of class imbalance through controlled resampling. 
Our results reveal that while IDCS classifiers improve cost-efficiency, they produce significantly less stable explanations compared to traditional models, particularly as class imbalance increases, highlighting a critical trade-off between cost optimization and interpretability in credit scoring. 
Amid increasing regulatory scrutiny on explainability, this research underscores the pressing need to address stability issues in IDCS classifiers to ensure that their cost advantages are not undermined by unstable or untrustworthy explanations.
\end{abstract}

% keywords can be removed
\keywords{Analytics \and Cost-sensitive learning \and Credit scoring \and Explainable AI \and Explanation stability}

% Main text
\section{Introduction}
\label{intro}

In credit scoring, the use of machine learning techniques to support or automate decision-making has become increasingly prevalent \citep{petrides2022cost}. 
While numerous advanced credit scoring models have been proposed to enhance discriminatory power \citep{GUNNARSSON2021292,lessmann2015benchmarking}, many overlook the financial consequences of misclassifying good or bad credits, especially when misclassification costs are asymmetrical and vary by loan.
As highlighted by \cite{DOUMPOS20231}, optimizing for financial outcomes, rather than solely statistical performance, is crucial for real-world relevance and a key research direction for artificial intelligence in banking.

In response to this challenge, several instance-dependent cost-sensitive (IDCS) classifiers have been developed to incorporate loan-specific misclassification costs directly into the model's loss function \citep{bahnsen2014example,bahnsen2015ensemble,bahnsen2015example,HOPPNER2022291,zelenkov2019example,vanderschueren2022instance}.
Empirical studies have shown that IDCS classifiers are more cost-effective compared to other cost-sensitive methods \citep{vanderschueren2022instance,petrides2022cost}, while their theoretical foundations have been further developed by \citet{CRELLA2025128875}.
As research into the financial benefits of these models advances, they show significant potential for future adoption by banks.

However, a limitation of these IDCS models is that they function as ``black-box'' algorithms, complicating the validation and understanding of their predictions. 
This critical trade-off between enhancing performance and maintaining interpretability has garnered significant attention in both academia \citep{bucker2022transparency,DUMITRESCU20221178} and financial regulation. 
For example, GDPR stipulates the `right to explanation' \citep{goodman2017ean} for credit scoring models,  the \cite{eba2021ai} recognized the importance of explainability, and credit scoring models are classified as ``high-risk AI systems'' in the proposed EU AI Act \citep{eu_ai_act_2024}.

To address these concerns, eXplainable AI (XAI) techniques like Local Interpretable Model-agnostic Explanations (LIME) \citep{lime} and SHapley Additive exPlanations (SHAP) \citep{lundberg2017unified} have emerged, with SHAP being particularly prominent in practice \citep{eba_dp_2023}. 
These explanation models offer post-hoc explanations of opaque models' predictions while preserving predictive performance.
However, to be truly interpretable, a model's explanations must be robust—meaning similar inputs should yield consistent explanations \citep{alvarez2018robustness}. 
Unstable explanations not only compromise the utility of XAI techniques but also expose financial institutions to legal and reputational risks \citep{goodman2017ean}, and can confuse customers trying to optimize their credit profiles.

Some studies have therefore explored XAI stability in credit scoring, but none address IDCS classifiers. 
\citet{visani2022robustness} introduced metrics to assess SHAP and LIME stability, while \citet{chen2024interpretable} analyzed the effects of class imbalance on explanation stability. 
Given that popular XAI techniques like SHAP and LIME rely on raw model predictions, the integration of instance-dependent costs into the loss function may fundamentally alter both the nature and stability of these explanations. 

This way, our study aligns with the XAIOR framework proposed by \cite{DEBOCK2024249}, which places cost-sensitive and profit-driven methods within the Performance Analytics (PA) dimension of risk assessment. 
In contrast, post-hoc interpretability techniques like SHAP and LIME are positioned as contributors to Attributable Analytics (AA) by making models interpretable, justifiable, and actionable. 
Our research advances the XAIOR agenda by exploring the trade-off between improving PA with IDCS methods and maintaining explanation stability, crucial for the AA dimension. 
We show that while IDCS models enhance performance, they can compromise explanation stability, a challenge that must be addressed before practical deployment.

We propose a two-faceted methodology evaluating prevalent IDCS classifiers in terms of costs and discriminatory power, and assessing the stability of SHAP and LIME when applied to these classifiers. 
We conduct a comparative analysis of traditional machine learning techniques  (i.e., XGBoost, logistic regression, random forest, and neural networks) alongside their IDCS variants using four open-source credit scoring datasets. 
To evaluate model performance, we report a mix of traditional and cost-sensitive metrics, being the Area Under the Curve (AUC), Average Precision (AP), Brier score, savings, and the relative Average Expected Cost (relAEC)---a novel, dimensionless adaptation of the Average Expected Cost (AEC) metric designed for cross-dataset comparisons.
To evaluate explanation stability, we measure the Coefficient of Variation (CoV) and the Sequential Rank Agreement (SRA) \citep{elstromSRA} of feature importances generated by both traditional and IDCS models. 
Following \cite{chen2024interpretable},  we also incorporate the additional effects of class imbalance through a controlled resampling procedure.

This work makes several key contributions to literature.
\begin{itemize}
	\item We are the first to assess the effect of using an IDCS classifier on the stability of SHAP and LIME explanations.
    \item We further expand the scope of explanation stability research beyond random forest and XGBoost \citep{chen2024interpretable} by including other popular models, such as logistic regression and neural networks \citep{lessmann2015benchmarking}.
	\item We adopt a dual focus on both model performance and model explanations, which is unique in the context of IDCS credit scoring.
	\item We introduce the relAEC metric, which offers a more meaningful comparison of cost savings across datasets than the AEC metric due to its relative nature.
	\item We provide the most complete assessment of classifiers with IDCS loss functions to date, by including all models in \cite{vanderschueren2022instance} and the random forest variant from \cite{bahnsen2015ensemble}.
\end{itemize}

The remainder of this study is organized as follows. 
Section \ref{sec:literature_review} reviews the relevant literature on IDCS credit scoring and the application of XAI in this domain, highlighting where the current research falls short. 
Section \ref{sec:methodology} outlines experimental framework, with the results presented in Section \ref{sec:experimental_results}. 
Section \ref{sec:discussion} discusses the findings, and Section \ref{sec:conclusions} draws conclusions and proposes avenues for future research.

\section{Related work}
\label{sec:literature_review}

Cost-sensitive learning incorporates the costs of correct and incorrect predictions into the decision-making process, aiming to predict the class that minimizes the expected cost by leveraging the conditional probability of each class for a given instance \citep{elkan2001}.
Cost-sensitive methods are classified based on whether they integrate costs at the class or instance level, and whether they do so directly or indirectly. 
While early research focused on class-dependent costs, \citet{vanderschueren2022instance} showed that instance-dependent costs---i.e. costs that are not fixed per class, but can vary per instance---yield superior cost savings across several tasks including credit scoring.

Direct methods incorporate costs directly into the classifier's loss function, while indirect methods or ``wrappers'' make cost-insensitive algorithms cost-sensitive by either pre-processing the training data or post-processing model outputs to account for costs. 
Since SHAP and LIME explanations are based on raw model outputs, indirect methods have no effect on them.
Therefore, this study focuses exclusively on direct IDCS classifiers that explicitly optimize for costs, potentially impacting both the nature and stability of these explanations.

The foundational work on IDCS classifiers was established by \citet{bahnsen2014example} with an IDCS logistic regression model, later extended to decision trees and tree ensemble methods \citep{bahnsen2015ensemble, bahnsen2015example}.
From here, \citet{zelenkov2019example} further expanded the approach to gradient-based ensembles, and more recent methods leveraging boosting frameworks such as XGBoost \citep{chen2016xgboost} and LightGBM \citep{ke2017lightgbm} have gained popularity due to their flexible loss functions. 
IDCS variants of these models have been applied in customer churn \citep{janssens2022b2boost}, fraud detection \citep{HOPPNER2022291}, forecasting \citep{haar2023} and credit scoring \citep{vanderschueren2022instance,vanderschueren2022predict}, where IDCS neural networks were also introduced.
Notably, in peer-to-peer (P2P) lending, a closely related field to traditional credit scoring, direct IDCS models have also gained popularity.
\citet{LI2021106963} applied IDCS LightGBM for early default detection, \citet{ARIZAGARZON2024101428} integrated instance-dependent profits into XGBoost to predict small-business defaults in P2P lending, and \citet{WU202274} introduced the Cost Sensitive Loan Evaluation (COSLE) framework, which optimizes decision trees and random forests using an IDCS Gini index. 

Beyond model development, \citet{petrides2022cost} benchmark the effectiveness of IDCS methods alongside other popular cost-sensitive methods in credit scoring and \citet{martin2025novel} propose an alternative instance-dependent cost matrix for credit scoring, aiming to improve upon the established approaches of \citet{bahnsen2015example} and \citet{VERBRAKEN2014505} that have become the standards.
Furthermore, \citet{CRELLA2025128875} have established a solid theoretical foundation for IDCS loss functions, reinforcing their empirical success.
Their study proves the consistency and asymptotic normality of IDCS estimators and is conducted in collaboration with a Spanish bank, highlighting growing interest in these methods by financial institutions.

Despite this increasing interest in IDCS classifiers, their impact on XAI techniques remains unexplored.
In credit scoring, two studies have examined XAI stability, though outside a cost-sensitive context.
\citet{visani2022robustness} introduced the Variable Stability Index (VSI) and Coefficient Stability Index (CSI) to assess LIME’s instability due to sampling randomness.
Building on this, \citet{chen2024interpretable} evaluated SHAP and LIME stability under varying class imbalances by measuring VSI, CSI, Coefficient of Variation (CoV), and Sequential Rank Agreement (SRA) \citep{elstromSRA}, finding that greater class imbalance leads to less stable explanations. 
Table \ref{tab:literature_review} summarizes the literature on IDCS learning and XAI in credit scoring, emphasizing the gap between these areas.

\begin{table}[ht]
	\newcommand{\X}{\checkmark}
	\footnotesize
    \setlength{\tabcolsep}{2pt}
	\renewcommand{\arraystretch}{0.7} %
	\caption{\textbf{Contributions to IDCS credit scoring literature.} Studies included either introduce new IDCS model variants, benchmark their performance or examine XAI stability in credit scoring.}   	
    \begin{tabularx}{\textwidth}{@{} l *{5}{>{\centering\arraybackslash}X} *{3}{>{\centering\arraybackslash}X} @{}}
		\toprule
        \multirow{2}{*}{\textbf{Studies}} &
        \multirow{2}{*}{\textbf{IDCS}} &
		\multicolumn{4}{c}{\textbf{Models}} & \multicolumn{3}{c@{}}{\textbf{XAI}} \\
		\cmidrule(lr){3-6} 
		\cmidrule(lr){7-9}
		& & \textbf{LR} & \textbf{DT/RF} & \textbf{Boost} & \textbf{NN} &  \textbf{SHAP} &  \textbf{LIME} & \textbf{Stability} \\
		\midrule
		\cite{bahnsen2014example}          & \X & \X &    &    &    &     &     &    \\
		\cite{bahnsen2015ensemble}         & \X &    & \X &    &    &     &     &    \\
		\cite{bahnsen2015example}          & \X &    & \X &    &    &     &     &    \\
		\cite{zelenkov2019example}         & \X &    &    & \X &    &     &     &    \\
		\cite{vanderschueren2022instance}  & \X & \X &    & \X & \X &     &     &    \\
		\cite{vanderschueren2022predict}   & \X & \X &    & \X & \X &     &     &    \\
		\cite{petrides2022cost}		   	   & \X & \X & \X & \X &    &     &     &    \\
		\cite{visani2022robustness}        &    & \X &    & \X &    &     & \X  & \X \\		
		\cite{de2023robust}                & \X & \X &    &    &    &     &     & 	  \\
		\cite{chen2024interpretable}       &    &    & \X & \X &    & \X  & \X  & \X \\
		\midrule
		\textbf{This study}                & \X & \X & \X & \X & \X & \X  & \X  & \X \\
		\bottomrule
	\end{tabularx}
	\label{tab:literature_review}
\end{table}

This table highlights critical gaps in the literature, which our study addresses in several ways.
Our primary contribution is pioneering an analaysis of how using an IDCS classifier influences the stability of XAI techniques, focusing on SHAP \& LIME.
Previous studies have primarily focused on either the robustness of IDCS classifiers' decision boundaries \citep{de2023robust} or the stability of XAI techniques for cost-insensitive classifiers \citep{chen2024interpretable}.
However, IDCS classifiers modify the learning process by embedding instance-dependent costs into the loss function, while XAI methods like SHAP and LIME generate explanations based on the model’s outputs.
To date, the effect of using an IDCS loss function on the consistency of the produced explanations remains unexplored.
Secondly, we expand the scope of models used in explanation stability studies by including popular credit scoring models like logistic regression and neural networks \citep{lessmann2015benchmarking}, whereas \cite{chen2024interpretable} focused solely on random forest and XGBoost.
Third, a dual focus is adopted on both model performance and model explanations, representing a novel approach in IDCS credit scoring.
Fourth, we introduce the relAEC metric, which facilitates comparisons of cost savings across datasets due to its relative nature, thereby enhancing the standard AEC metric.
Finally, we present the most comprehensive comparative analysis of IDCS classifiers to date, incorporating models from \cite{vanderschueren2022instance} and the random forest implementation from \cite{bahnsen2015ensemble}.

\section{Methodology}
\label{sec:methodology}

\begin{figure}[ht]
	\centering
	\includegraphics[width=.85\textwidth]{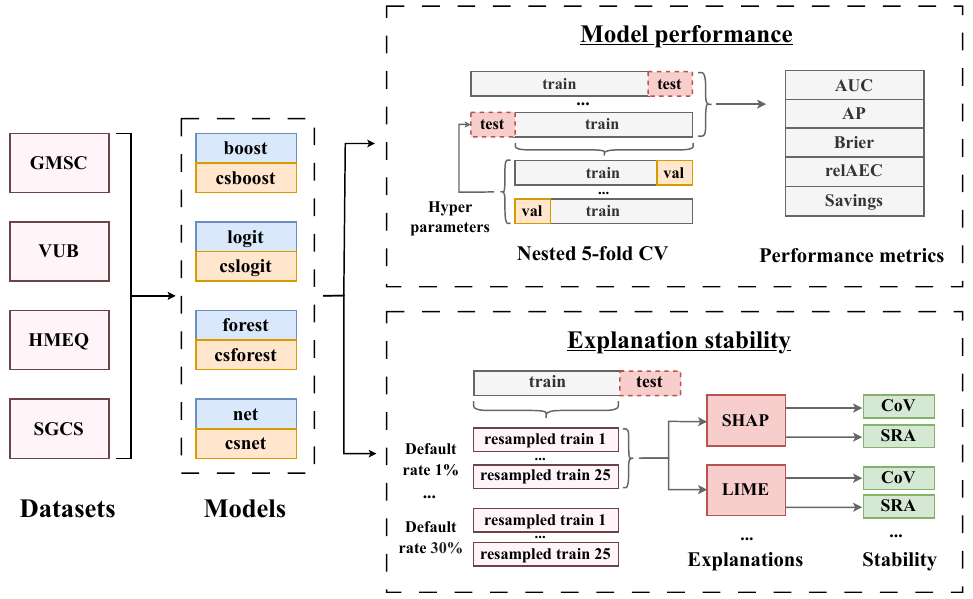}
	\caption{Proposed experimental setup}
	\label{fig:full_methodology}
\end{figure}

We adopt a dual approach to assess both the performance of IDCS classifiers and the stability of their explanations, as illustrated in Figure \ref{fig:full_methodology}.

The experiment is performed on four widely-used credit scoring datasets varying in size, number and types of variables, and default rates. The models considered are \textit{boost} (XGBoost), \textit{logit} (logistic regression), \textit{forest} (random forest), \textit{net} (neural networks) and their IDCS variants, \textit{csboost}, \textit{cslogit}, \textit{csforest}, and \textit{csnet}.
Model performance is measured using nested 5-fold cross-validation, with hyperparameters tuned in the inner loop and performance metrics cross-validated in the outer loop.
For explanation stability, we perform controlled resampling, following \cite{chen2024interpretable}, adjusting training set default rates between 1\% and 30\% while keeping a constant sample size.
In 25 random resampling iterations per default rate, we train the models on the resampled dataset and compute SHAP and LIME importances on the test set, evaluating stability between the feature importance lists via the coefficient of variation (CoV) and Sequential Rank Agreement (SRA) \citep{elstromSRA}.
Each step of our proposed methodology is presented in the following sections. 
%The implementation, fully developed in Python, is publicly accessible at github-link

\subsection{Data}
\label{sec:data}

The four publicly available credit scoring datasets are selected because of their popularity in research \citep{vanderschueren2022predict,devos2018profit,lessmann2015benchmarking} and offer a representative set for the field of credit scoring. 
The selection of these datasets was guided by IDCS learning's requirement for a variable related to the requested loan amount, which is often missing in publicly available datasets \citep{martin2025novel}.
The selected set encompasses all relevant datasets used in recent IDCS credit scoring studies.
The datasets vary not only in terms of sample sizes but also in their degree of class imbalance, as summarized in Table \ref{tab:datasets}.
In this context, the target variable signifies default or severe delinquency, with explanatory variables encompassing loan attributes, prior customer conduct, socio-economic factors, and other relevant indicators. 
For an overview of the variables per dataset we refer to the supplementary material. 

\begin{table}[ht]
	\footnotesize
	\centering
	\renewcommand{\arraystretch}{0.75} %
	\caption[Overview of used credit scoring datasets]{\textbf{Dataset overview.} Abbreviation (Abbr.), size (N), dimensionality (D) and default rate (\% Pos)}
	\begin{tabular}{lcccc}
		\toprule
		\textbf{Dataset} & \textbf{Abbr.} & \textbf{N} & \textbf{D} & \textbf{\% Pos} \\
		\midrule
		\textbf{Kaggle Give Me Some Credit} \citep{GiveMeSomeCredit} & GMSC & 112,915 & 10 & 6.74\\
		\textbf{VUB Credit Scoring} \citep{petrides2022cost} & VUB & 18,667 & 20 & 16.95 \\
		\textbf{Home Equity} \citep{baesens2016credit} & HMEQ & 5,960 & 12 & 19.95 \\
		\textbf{South-German Credit Scoring} \citep{south_german_credit_2019} & SGCS & 1,000 & 20 & 30 \\
		\bottomrule
	\end{tabular}
	\label{tab:datasets}		
\end{table}
\subsection{Models}
\label{sec:models}

We conduct our experimental analysis using various models from the existing IDCS credit scoring literature, being \textit{csboost}, \textit{cslogit}, \textit{csnet} and \textit{csforest}. 
The first two are the IDCS versions of XGBoost and logistic regression developed by \cite{HOPPNER2022291} in R. 
We build upon the Python implementations provided by \cite{vanderschueren2022instance} and add their IDCS neural network, \textit{csnet}.
Additionally, we include the IDCS random forest model proposed by \cite{bahnsen2015ensemble}, referred to as \textit{csforest} in this study.
Important to note is that, while csboost, cslogit and csnet directly minimize the Average Expected Cost (AEC) metric, this is not the case for csforest. 
Instead, the IDCS random forest is built upon IDCS decision trees that use the cost of classifying all instances in a node as 1 or 0 as splitting criterium, and the decrease in total cost of the algorithm as the gain of using each split.
Next to the IDCS models, we perform the experiment for the traditional cost-insensitive versions, minimizing cross-entropy (CE), hereafter referred to as \textit{boost}, \textit{logit}, \textit{net} and \textit{forest}.

\subsection{Instance-dependent credit scoring cost matrix}
\label{sec:costimplementation}

To incorporate costs into the cost-sensitive credit scoring models, we formulate the cost matrix for each instance $i$ as a function of the loan amount $A_i$ and several (fixed) parameters, following the approach by \cite{bahnsen2014example}.
This results in the cost matrix in Table \ref{tab:cost-matrix-credit}. 

\begin{table}[ht]
	\caption{\textbf{Instance-dependent credit scoring cost matrix.} This matrix assigns a specific cost to each potential outcome based on both the actual and predicted class.}
	\centering
	\footnotesize % Reduce text size
	\setlength{\tabcolsep}{6pt}
	\begin{tabular}{ll|cc}
        \multicolumn{2}{c}{}&\multicolumn{2}{c}{\textbf{Actual}}\\
        \multicolumn{2}{c|}{}&\textbf{No default}& \textbf{Default}\\
   		\cline{2-4}
        \multirow{2}{*}{\parbox{1.8cm}{\centering \textbf{Predicted}}}& \textbf{No default} & $0$ & $C_{i}(0|1) = A_i\cdot LGD$ \\
         & \textbf{Default}  & $C_{i}(1|0) = r_i + C_{alt}$ & $0$\\ 
   	\end{tabular}
	\label{tab:cost-matrix-credit}
\end{table}

In this study, we adhere to the convention where the positive class (1) denotes the minority class, being bad credit, and the negative class (0) signifies the good credit \citep{elkan2001}. 
The costs associated with correct classifications, $C_{i}(1|1)$ and $C_{i}(0|0)$, are set to zero for each loan applicant $i$ following the general convention in literature \citep{elkan2001,bahnsen2014example}. 
To determine the cost of incorrectly granting a loan to a defaulter, $C_{i}(0|1)$, we assume that the bank's losses are directly proportional to the loan amount $A_i$, using a constant Loss Given Default (LGD) parameter of $0.75$ \citep{gcd_reports}. 
This approach is chosen over more sophisticated methods, such as an instance-based LGD through an additional model, as these have demonstrated limited performance \citep{petrides2022cost}. 

The cost of not granting a loan to a trustworthy borrower, $C_{i}(1|0)$, includes two components \citep{bahnsen2014example}: the lost revenue from denying customer $i$ a loan ($r_i$), and the expected cost of lending to the bank's average alternative customer ($C_{alt}$).
The second term arises from the assumption that the financial institution will not keep the ungranted funds idle, but will instead find an alternative debtor. 
This alternative borrower is assumed to borrow an average loan amount $\overline{A}$, yielding an average profit $\overline{r}$, and repays this debt with a probability of $\pi_0$ (prior negative rate), or defaults with probability $\pi_1$ (prior positive rate). The average loan amount $\overline{A}$ is derived solely using costs within the training set to prevent data leakage.
With these parameters, the following can be derived:

\begin{equation}\label{eq:FP}
	C_{i}(1|0) \; = \; r_i + C_{alt} \; = \;  r_i + (-\overline{r} \cdot \pi_0 + \overline{A}\cdot LGD \cdot \pi_1)
\end{equation}

This means that the cost of a false positive is the lost revenue from not lending to customer $i$, minus the expected profits from lending to an average alternative good borrower, plus the expected losses from lending to an average defaulting borrower. 
Note that the prior rates $\pi_0$ and $\pi_1$ are fixed during the resampling procedure in the explanation stability experiment to ensure comparability across resampled default rates.
For instance, if the training set is resampled to 1\% versus 10\% defaulters, calculating the expected cost for an alternative customer using the resampled rates would yield significantly different costs for granting the alternative loan ($C_{alt}$)—the former being tenfold lower.
This could lead to overly stringent models when the training set is resampled to higher class imbalance levels, as the alternative customer's risk gets underestimated. 
Instead, following \cite{bahnsen2014example}, we adhere to real-world practices where a bank determines the alternative customer's chance of default using its actual historical default rates.

\subsection{Average Expected Cost as cost-sensitive loss function}
\label{sec:aec}

Given the (mis)classification costs in Table \ref{tab:cost-matrix-credit}, minimizing losses equals predicting each instance to have the class leading to the smallest expected loss \citep{elkan2001}. 
To obtain these expected losses, machine learning algorithms compute the conditional probability of each class given the instance. 
More specifically, given $P$ explanatory variables $\textbf{X} = (X_1, . . . , X_P)$ and binary response variable $Y \in \{0, 1\}$, a classification algorithm models the conditional expected value of Y on the basis of observed variables $\textbf{x} \in \textbf{X}$ \citep{HOPPNER2022291}:

\begin{equation}\label{eq:condexpvalue}
	s: \textbf{X} \rightarrow [0,1]: \textit{x} \rightarrow	 s(\textbf{x}) = \mathbb{E}(Y \mid \textbf{x}) = P(Y = 1 \mid \textbf{x})
\end{equation}

This results in a score $s(\textbf{x}_i) \in [0,1]$ for every loan applicant, indicating the propensity of default in this case where the defaulting class is the positive class. 
If each observation has a cost matrix $\textbf{C}_i$, the corresponding empirical risk is the Average Expected Cost (AEC):

\begin{equation}\label{eq:aec}
	\begin{aligned}
		AEC(y_i, s(\textbf{x}_i), \textbf{C}_i) =  &\, y_i \Bigl(s(\textbf{x}_i)\cdot C_i(1|1) + (1 - s(\textbf{x}_i))\cdot C_i(0|1) \Bigr) \\
		&\, + (1 - y_i) \Bigl( s(\textbf{x}_i)\cdot C_i(1|0) + (1 - s(\textbf{x}_i))\cdot C_i(0|0)\Bigr)
	\end{aligned}
\end{equation}

This metric is threshold-independent, as it relies on the estimated probabilities $s(\textbf{x}_i)$ instead of the predicted classes $\hat{y}_i$. 
Using the cost matrix and parameters defined in Section \ref{sec:costimplementation}, in which $C_i(0|0)$ and $C_i(1|1)$ are set to 0, we obtain:

\begin{equation}\label{eq:aec_simplified}
	\begin{aligned}
		AEC(y_i, s(\textbf{x}_i),A_i) =  &\, y_i \Bigl((1 - s(\textbf{x}_i))\cdot A_i\cdot LGD\Bigr) \\
		&\, + (1 - y_i) \Bigl(s(\textbf{x}_i)\cdot (r_i + (-\overline{r}\cdot \pi_0 + \overline{A}\cdot LGD\cdot \pi_1))\Bigr)
	\end{aligned}
\end{equation}

AEC ranges from 0 to $+\infty$ and should be minimized for cost-effective learning. 
Since its scale depends on the instance-dependent costs in the data, it varies across datasets, complicating performance comparisons.
To address this, we introduce the relative AEC (relAEC), which normalizes AEC against a baseline of using no model, thereby making it dimensionless. 
The baseline predicts a probability estimate $s(\textbf{x}_i)$ equal to the prior default rate $\pi_1$ for all observations to compute the AEC.
For example, in the GMSC dataset (Section \ref{sec:data}), the baseline predicts a default probability of 0.0674 for every instance. Mathematically, we can define relAEC as:

\begin{equation}\label{eq:relaec}
	\begin{aligned}
		relAEC(y_i, s(\textbf{x}_i),A_i) =  1 - \frac{AEC(y_i, s(\textbf{x}_i),A_i)}{AEC(y_i, \pi_1,A_i)}
	\end{aligned}
\end{equation}

This yields a value between -$\infty$ and 1 given non-negative costs and zero profits for correct classifications, as in our setup. 
A relAEC of 1 indicates a perfect model with an AEC of 0, where all defaults are perfectly predicted (probability of 1) and non-defaults receive a probability of 0.
A relAEC of 0 then signifies performance equal to the prior probability baseline, values between 0 and 1 show better performance than the baseline, and negative values reflect worse performance.
This interpretation closely aligns with the \textit{savings} metric from \cite{bahnsen2014example}.

\subsection{Explanation techniques}

This study examines the stability of Shapley Additive Explanations (SHAP) \citep{lundberg2017unified} and Local Interpretable Model-agnostic Explanations (LIME) \citep{lime}, two widely used post hoc explainability techniques.
SHAP assigns contributions to each feature using the Shapley Value \citep{shapley1953value} from cooperative game theory, ensuring a theoretically sound attribution of feature importance.
LIME, in contrast, explains individual predictions by perturbing input features and training a locally interpretable model (e.g., ridge regression) weighted by proximity to the original instance.

SHAP and LIME were selected for three key reasons. 
First, they provide local explanations, essential for financial decision-making to ensure fairness, compliance, and transparency \citep{goodman2017ean}.
This is important, as both credit officers and customers are particularly interested in the decomposition of individual credit decisions as obtained by local methods \citep{bucker2022transparency}.
Second, their widespread adoption in both research and financial institutions supports their relevance; a recent industry survey \citep{eba_dp_2023} found Shapley values to be the most commonly used interpretability tool in financial applications, with 40\% of the 60\% of institutions utilizing interpretability techniques relying on them.
Additionally, SHAP and LIME are by far the most dominant post-hoc explainability methods in recent credit scoring research, as summarized by \citet{chen2024interpretable}.
Third, their model-agnostic nature ensures compatibility and direct comparability across different classifiers, making them ideal for this comparative study.

Post-hoc explanation techniques like SHAP and LIME have notable limitations, however. 
Prior research highlights stability issues for both methods under class imbalance \citep{chen2024interpretable}, with LIME also facing broader stability and fidelity concerns \citep{janssens2023360, visani2022robustness}. 
Both methods are also vulnerable to adversarial manipulation, as demonstrated by \cite{slack2020fooling}, who show how a scaffolding technique can hide biases, such as discrimination against female applicants, from the explanations.
There is also a broader limitation concerning explainability, which involves a trade-off between offering simpler, more interpretable explanations and providing more complex yet accurate ones \citep{szepannek2024can}.
Therefore, these techniques should be critically assessed, even for traditional models.
Given their widespread adoption, however, the stability of SHAP and LIME remains a critical concern—a challenge our study extends to the context of IDCS classifiers.

This study uses \textit{PermutationExplainer} \citep{strumbelj2014explaining} for SHAP, iterating through feature permutations in forward and reverse directions to compute the SHAP values. 
For LIME, we employ \textit{LimeTabularExplainer} \citep{lime} with ridge regression and an exponential kernel based on Euclidean distance to weight feature proximity.
Feature importances are derived from the SHAP values and LIME coefficients, with higher absolute values indicating greater influence. Sorting them provides a feature importance ranking.

\subsection{Stability metrics}
To assess the stability of SHAP \& LIME, we compute two stability metrics: the Coefficient of Variation (CoV) and the Sequential Rank Agreement (SRA) \citep{elstromSRA}.
The CoV evaluates the stability of feature importance values, while the SRA assesses the stability of their rankings. 
Both stability metrics are hereafter referred to using $\delta^{CoV}$ and $\delta^{SRA}$ respectively.
The CoV of an instance $i$ ($\delta^{CoV}_{i}$) is computed as the average CoV across all P features' importances for that instance:

\begin{equation}
	\delta^{CoV}_{i} = \frac{1}{P} \sum_{p} \delta^{CoV}_{ip} = \frac{1}{P} \sum_{p} \frac{\sigma_{ip}}{\mu_{ip}}
\end{equation}

where \( \sigma_{ip} \) is the standard deviation of the importance of feature $p$ and \( \mu_{ip} \) is its mean. 

\begin{figure}[ht]
	\centering
	\includegraphics[width=0.55\textwidth]{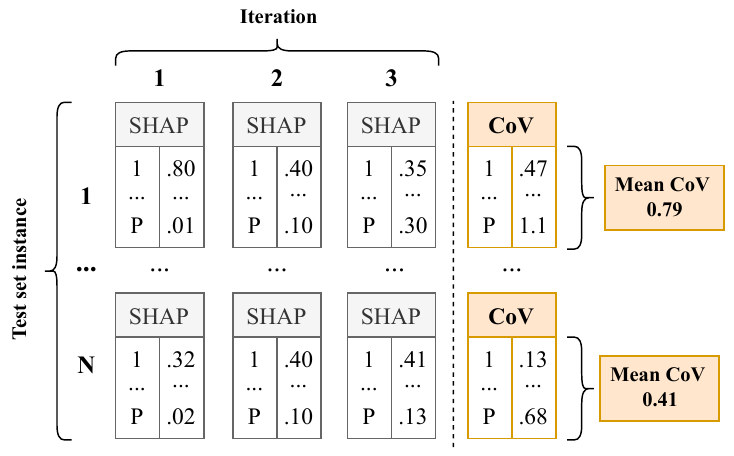}
	\caption[Example computation of SHAP CoV]{\textbf{Example SHAP CoV computation across 3 iterations.} For each feature, we first compute the CoV value across iterations, after which we average the CoV of all the features to get an estimate per instance.}
	\label{fig:covstability}
\end{figure}

This metric quantifies the stability of feature importance values across iterations.
Due to its dimensionless nature, it can be easily compared between standard and IDCS model variants, as well as across datasets.
Figure \ref{fig:covstability} illustrates the calculation of the SHAP CoV for each instance across three iterations at a given imbalance level.

While the CoV assesses the stability of feature importance \textit{values}, the Sequential Rank Agreement (SRA) \citep{elstromSRA} evaluates the stability of feature importance \textit{rankings}. 
SRA quantifies the agreement in rankings among two or more ordered lists, making it suitable for comparing the ranked feature importance lists produced by SHAP and LIME. 
An example of how feature importance lists are formatted for computing SRA is provided in Table \ref{tab:SRA}.

\begin{table}[ht]
	\footnotesize
	\centering
	\renewcommand{\arraystretch}{0.8} 
	\caption{\textbf{Example SRA calculation.} Three feature importance lists (a) with features ${A, B, C, D, E}$ are converted into ranking lists (b). The SRA is then computed for each depth (c).}
	\begin{subtable}{0.27\textwidth}
		\centering
		\label{subtab:a}
		\begin{tabularx}{\textwidth}{>{\centering\arraybackslash}X *{3}{>{\centering\arraybackslash}X}}
			\hline
			$\mathbf{p}$ & $\mathbf{\Phi_1}$ & $\mathbf{\Phi_2}$ & $\mathbf{\Phi_3}$ \\
			\hline
			\textbf{A} & .20 & .21 & .36 \\
			\textbf{B} & .84 & .22 & .28 \\
			\textbf{C} & .55 & .33 & .89 \\
			\textbf{D} & .93 & .65 & .77 \\
			\textbf{E} & .12 & .10 & .11 \\
			\hline
		\end{tabularx}
		\caption{Feature importance value lists}
	\end{subtable}
	\hfill
	\begin{subtable}{0.3\textwidth}
		\centering
		\label{subtab:b}
		\begin{tabularx}{\textwidth}{>{\centering\arraybackslash}X *{4}{>{\centering\arraybackslash}X}}
			\hline
			$\mathbf{p}$ & $\mathbf{R_1}$ & $\mathbf{R_2}$ & $\mathbf{R_3}$ & $\mathbf{\hat{A_L}}$ \\
			\hline
			\textbf{A} & 4 & 4 & 3 & .33 \\
			\textbf{B} & 2 & 3 & 4 & 1   \\
			\textbf{C} & 3 & 2 & 1 & 1   \\
			\textbf{D} & 1 & 1 & 2 & .33 \\
			\textbf{E} & 5 & 5 & 5 & 0   \\
			\hline
		\end{tabularx}
		\caption{Feature importance ranking lists}
	\end{subtable}
	\hfill
	\begin{subtable}{0.36\textwidth}
		\centering
		\label{subtab:c}
		\begin{tabularx}{\textwidth}{>{\centering\arraybackslash}X>{\centering\arraybackslash}p{2.5cm}>{\centering\arraybackslash}X}
			\hline
			\textbf{Depth} &\textbf{S(d)} & \textbf{SRA} \\
			\hline
			\textbf{1} &$\{C, D\}$			& .67 \\
			\textbf{2} &$\{B, C, D\}$		& .77 \\
			\textbf{3} &$\{A, B, C, D\}$	& .67 \\
			\textbf{4} &$\{A, B, C, D\}$	& .67 \\
			\textbf{5} &$\{A, B, C, D, E\}$	& .53 \\
			\hline
		\end{tabularx}
		\caption{Feature sets and SRA in function of depth}
	\end{subtable}
	\vspace{-10pt}
	\label{tab:SRA}
\end{table}

For each list in (a), the importances are converted to rankings, with 1 denoting the highest importance, resulting in the ranking lists displayed in panel (b). 
Empirically, the agreement among $L$ lists regarding the rank assigned to a feature $p$ can be measured by the variance in the rankings across these lists:

\begin{equation}
	\widehat{A_L}(p) = \frac{1}{L - 1} \sum_{i=1}^{L} (R_i(p) - \overline{R_L}(p))^2,
\end{equation}

where $\overline{R_L}(p) = \frac{1}{L} \sum_{i=1}^{L} R_i(p)$ is the average rank given to feature $p$ across all $L$ lists. 

The SRA metric can be applied at various \textit{depth} levels $d$, focusing on the ranking stability of a subset $S(d)$ of features that were ranked within the top $d$ positions in any of the lists, as illustrated in panel (c). 
For instance, with a depth of 3, only features that appear in the top 3 ranking of at least one list are considered, excluding feature $E$ in panel (c). 
Using a depth smaller than the full feature set is practical, as not all features may have a meaningful impact on the prediction outcome. 
Thus, variations in the rankings of less important features are considered trivial. 
The SRA among $L$ lists at a given depth $d$ ($\delta^{SRA}_{d}$) is then calculated as the pooled variance of the items in $S(d)$:

\begin{equation}
	\delta^{SRA}_{d} = \frac{1}{|S(d)|} \sum_{p \in S(d)} \frac{(L - 1) \widehat{A_L}(p)}{(L - 1)|S(d)|}.
\end{equation}

A value close to zero indicates that the lists agree on the rankings, whereas larger values suggest greater disagreement and in this case more instability. 
In our experiments, we use a depth of 10, meaning that we include a feature in the SRA calculation if it appears in the top 10 most important features in at least one of the lists. Just as shown for CoV in Figure \ref{fig:covstability}, the SRA is computed for every instance ($\delta^{SRA}_{i}$) as a measure of the stability of feature rankings values across iterations at a given default rate.

\subsection{Model performance: discriminative and cost}
\label{sec:performance_exp}

The first part of this study evaluates and compares the performance of the models specified in Section \ref{sec:models} using a nested 5-fold cross-validation approach. 
The procedure consists of two steps: the outer loop evaluates model performance, while the inner loop tunes hyperparameters.
This ensures performance evaluation is independent of hyperparameter tuning, reducing bias.
In each outer loop iteration, 20\% of the data is reserved for testing, with the remaining 80\% used for training. 
On each outer training fold, an inner loop performs a 5-fold CV grid search over candidate hyperparameters (listed in Table \ref{tab:hyperparameters}) to identify the optimal values. 
These hyperparameter ranges are set based on those used in prior IDCS studies \citep{vanderschueren2022predict,janssens2022b2boost}. 
Models in the outer loop are then trained with the best hyperparameters from the inner loop and evaluated on the outer test fold. 
The complete procedure is outlined in Algorithm \ref{algo:performance_evaluation}.

\alglanguage{pseudocode}
\begin{algorithm}[ht]
	\footnotesize
	\caption{Evaluating model performance}
	\setlength{\baselineskip}{12pt}
	\begin{algorithmic}
		\State{\textbf{Result}: evaluation metric matrix}
		\State {\textbf{Set} $models$ \textbf{To} $[boost,csboost,logit,cslogit,forest,csforest,net,csnet]$}
		\State Load data
		\State One-hot-encode categorical variables
		\State Split data into 5 stratified folds 
		\For{$i \in [1:5]$}	
		\State Test data = fold $i$
		\State Training data = all folds except $i$
		\State Initialize cost matrices based on training data
		\State Standardize numerical variables based on training data
		\For{$model \in [models]$}
		\State Tune for optimal hyperparameters $\theta^*$ using 5-fold CV on training set
		\State Train model on full training set using $\theta^*$
		\State Predict on fold $i$
		\State Compute AUC, AP, Brier Score, relAEC and savings for fold $i$
		\EndFor
		\EndFor
		\State Compute average and sdev of evaluation metrics over all folds
	\end{algorithmic}
	\label{algo:performance_evaluation}
\end{algorithm}	

\begin{table}[t]
	\centering
	\footnotesize
	\caption{Hyperparameter search space}
	\renewcommand{\arraystretch}{0.8} 
    \begin{tabularx}{\textwidth}{@{} >{\centering\arraybackslash}X >{\centering\arraybackslash}X >{\centering\arraybackslash}X @{}}
		\toprule
		\textbf{Model} & \textbf{Parameter} & \textbf{Candidate values} \\
		\midrule
		\textbf{(cs)boost} 		& Learning Rate & 0.3 \\
		& Min child weight & \{0,50\} \\
		& Max depth & \{1,3,5,7\} \\
		& colsample\_bytree & \{0.8, 1\} \\
		& $\gamma$ & \{0,5,10\} \\
		\textbf{(cs)logit}	& Penalty  & \{L1, L2\} \\
		& Inverse of regularization strength C & \parbox[t]{5cm}{\centering \setlength{\baselineskip}{12pt} \{0, 1e-4, 3e-4, 6e-4, 1e-3, 3e-3\\ 6e-3, 1e-2, 3e-2, 1e-1\}} \\
		\textbf{(cs)forest} & Max depth & \{-, 3, 5, 10\} \\
		& Number of estimators &\{20, 50, 100\} \\
		\textbf{(cs)net} 	& Learning Rate & 0.005  \\
		& Number of neurons & \{32,64,128\} \\
		\bottomrule
	\end{tabularx}
	\label{tab:hyperparameters}
\end{table}
The result is a dictionary containing averaged performance metrics for each model across folds.
For cost-insensitive metrics, the Area Under the Curve (AUC), Average Precision (AP), and Brier score are reported due to their popularity in credit scoring and their combined ability to cover all aspects of model performance \citep{lessmann2015benchmarking}.
The AUC measures the probability that a randomly chosen positive instance (default) scores higher than a negative one (non-default). 
AUC values range from 0 to 1, with 0.5 indicating that the model cannot discriminate between classes better than random guessing.
As argued by \cite{davis2006relationship}, the AUC metric can give overly optimistic results when the class distribution is skewed, prompting the inclusion of Average Precision (AP), which measures the area under the precision-recall curve and has a baseline equal to the positive class proportion \citep{saito2015precision}.
For the VUB dataset with a 16.95\% default rate, a random classifier achieves an AUC of 0.5 and an AP of 0.1695, while a perfect classifier reaches 1 for both metrics.
Lastly, the Brier score evaluates prediction accuracy by computing the mean squared error between predicted probabilities and binary outcomes, indicating how well-calibrated the model's probability estimates are \citep{lessmann2015benchmarking}. 
Its value ranges from 0 to 1, with lower scores indicating better calibration.
The null model Brier Score is obtained by consistently predicting the prior default rate for all instances.

In terms of cost-sensitive metrics, the relative AEC (relAEC) is evaluated as defined in Section \ref{sec:aec}, along with model savings \citep{bahnsen2014example}.
The savings metric is designed to measure the cost reduction achieved by using classifier $f$ on dataset $D$ compared to classifying all instances into a single class. 
To accomplish this, the costs of classifying all instances in the test set as credit-worthy ($C(0|D)$) or as a defaulter ($C(1|D)$) are computed. The baseline cost $C_{base}(D)$ is then defined as:

\begin{align}
	\label{eq:costempty}
	C_{\text{base}}(D) &= \min\{C(0|D), C(1|D)\} 
\end{align}

This is illustrated through the hypothetical cost matrices presented in Figure \ref{fig:savings_example}. 

\begin{figure}[ht]
	\centering
	\begin{subfigure}[b]{0.3\textwidth}
		\centering
		\footnotesize
		\begin{tabular}{ll|cc}
            \multicolumn{2}{c}{} & \multicolumn{2}{c}{\footnotesize \textbf{Actual}} \\
			\multicolumn{2}{c|}{} & \textbf{0} & \textbf{1} \\
			\cline{2-4}
            \multirow{2}{*}{\rotatebox{0}{\footnotesize \textbf{Predicted}}} & \textbf{0} & $0$ & $10$ \\
			& \textbf{1} & $3$ & $0$ \\
		\end{tabular}
		\caption{Default}		
	\end{subfigure}%
	\begin{subfigure}[b]{0.3\textwidth}
		\centering
		\footnotesize
		\begin{tabular}{ll|cc}
            \multicolumn{2}{c}{} & \multicolumn{2}{c}{\footnotesize \textbf{Actual}} \\
			\multicolumn{2}{c|}{} & \textbf{0} & \textbf{1} \\
			\cline{2-4}
            \multirow{2}{*}{\rotatebox{0}{\footnotesize \textbf{Predicted}}} & \textbf{0} & $0$ & $7$ \\
			& \textbf{1} & $1$ & $0$ \\
		\end{tabular}
		\caption{Non-default}
	\end{subfigure}%
	\begin{subfigure}[b]{0.3\textwidth}
		\centering
		\footnotesize
		\begin{tabular}{ll|cc}
            \multicolumn{2}{c}{} & \multicolumn{2}{c}{\footnotesize \textbf{Actual}} \\
			\multicolumn{2}{c|}{} & \textbf{0} & \textbf{1} \\
			\cline{2-4}
            \multirow{2}{*}{\rotatebox{0}{\footnotesize \textbf{Predicted}}} & \textbf{0} & $0$ & $6$ \\
			& \textbf{1} & $5$ & $0$ \\
		\end{tabular}
		\caption{Non-default}
	\end{subfigure}
	\caption[Baseline savings example]{Baseline savings example with one actual defaulter and 2 non-defaulters}
	\label{fig:savings_example}
\end{figure}

Assuming only instance (a) is an actual defaulter, $C(0|D)$ and $C(1|D)$ are computed by predicting either 0 or 1 for all cases respectively, yielding a baseline cost:

\begin{align}
	\label{eq:costempty_ex}
	C_{\text{base}}(D) &= \min\{(10+0+0), (0+1+5)\} = 6
\end{align}

This baseline cost serves as a normalization factor when evaluating the actual costs incurred by using the model, denoted as $C_{f}(D)$, allowing the calculation of the model's savings:

\begin{align}
	\label{eq:savings}
	\text{Savings}_{f}(D) &= 1 - \frac{C_{f}(D)}{C_{\text{base}}(D)}
\end{align}

Its value ranges between -$\infty$ and 1, similarly to the proposed relAEC metric. 
A score of 1 indicates a perfect model that correctly classifies each instance, while a score of 0 signifies performance equivalent to classifying all instances into a single class. 
Values between 0 and 1 reflect improvements over this baseline classification, whereas negative values indicate performance that is worse than always predicting the same class.

\subsection{Explanation stability}
\label{sec:explanation_evaluation}

The effect of using an IDCS classifier on the stability of SHAP and LIME is assessed following the procedure in Algorithm \ref{algo:explanation_evaluation}. 
Let $\pi \in \Pi$ be the default rates used to resample the training set, simulating various class imbalances, $i \in N$ the test set observations for stability assessment, and $j \in J$ a set of iterations. 
The SHAP and LIME feature importance values generated at imbalance level $\pi$ for instance $i$ in iteration $j$ are represented as $\Phi^{SHAP}_{\pi ij}$ and $\Phi^{LIME}_{\pi ij}$. 
These values are then used to compute stability metrics $\delta^{CoV}_{\pi i}$ and $\delta^{SRA}_{\pi i}$ for each test observation $i$ at imbalance level $\pi$. 

\alglanguage{pseudocode}
\begin{algorithm}[ht]
	\footnotesize
	\setlength{\baselineskip}{12pt}
	\begin{algorithmic}\caption{Evaluating explanation stability}
        \State \textbf{Set} $models$ \textbf{To} $[\texttt{boost}, \texttt{csboost}, \texttt{logit}, \texttt{cslogit}, \texttt{forest}, \texttt{csforest}, \texttt{net}, \texttt{csnet}]$
		\State{\textbf{Set} $\Pi$ \textbf{To} $[0.01,0.03,0.05,0.1,0.15,0.2,0.25,0.3]$}
		\State {\textbf{Set} $J$ \textbf{To} 25}
		\State {Split off a stratified test set $N$ based on the original imbalance level}
		\For {$model$ \textbf{in} $models$}
		\For {$\pi$ \textbf{in} $\Pi$}
		\State {\textbf{Set} $\theta^*$ \textbf{To} $\{\}$}
		\State {\textbf{Set} $tuning\_round$ \textbf{To} $True$}
		\For {$j \in J$}
		\State {Resample training set to imbalance level $\pi$}
		\State {Preprocess training set and test set}
		\If{$tuning\_round$} 
		\State {Perform 5-fold CV and store optimal hyperparameters $\theta^*$}		
		\EndIf
		\State {\textbf{Set} $tuning\_round$ \textbf{To} $False$}
		\State{Train model using hyperparameters $\theta^*$}
		\State{Fit SHAP and LIME explainers on training set}
		\State{Compute and store $\Phi^{SHAP}_{\pi ij}$ and $\Phi^{LIME}_{\pi ij}$ for each test set observation $i$}
		\EndFor
		\State{Compute and store stability metrics $\delta_{\pi i}$}
		\EndFor
		\EndFor
		\label{algo:explanation_evaluation}
	\end{algorithmic}
\end{algorithm}	

For each dataset, a stratified test set $N$ is partitioned, preserving the original default rate for representativeness, with a fixed size of 300 observations.
The size of the test set balances robust estimates of explanation stability with computational feasibility.
While \cite{chen2024interpretable} used 200 observations across five repetitions for a total of 1000, our study employs a slightly smaller set.
This choice accommodates the inclusion of more models and datasets, ensuring reliable results while managing computational costs, particularly given the demands of repeatedly generating SHAP and LIME values.

Training data is resampled to levels between 1\% and 30\%. 
Resampling beyond a $30/70$ ratio is deemed trivial, as studies suggest limited benefits beyond this threshold \citep{estabrooks2004multiple} and ratios of $20/80$ to $30/70$ are generally considered effective \citep{kamalov2022partial,baesens2015fraud}.
All resampled datasets have a fixed size of $\frac{\text{\# defaults}}{0.3}$, the highest fixed size that retains all minority samples while adhering to the maximum resampling default rate of 30\%.
This approach eliminates the impact of varying sample sizes on our results.

For each default rate, models are trained on the resampled training set across $J$ iterations.
The first iteration at each default rate is a tuning round in which hyperparameters are tuned through 5-fold cross-validation to identify optimal parameters $\theta^*$.
The models are then trained using the identified $\theta^*$ on the resampled training set, generating SHAP and LIME feature importance lists $\Phi^{SHAP}_{\pi ij}$ and $\Phi^{LIME}_{\pi ij}$ for each test instance $i$. 
Once all iterations at a given imbalance level are completed, these feature importance values are used to compute stability measures $\delta_{\pi i}^{CoV}$ and $\delta_{\pi i}^{SRA}$ across iterations.
A total of 25 iterations are performed to ensure stable estimates of the stability metrics.

\section{Experimental Results}
\label{sec:experimental_results}

\subsection{Model performance: discriminative and cost}
\label{sec:results_modelperformance}

Before delving into the main results of this study, we begin by presenting a comparative analysis of the performance across all models included in our evaluation.
This section outlines the out-of-sample performance estimates derived from the nested 5-fold cross-validation described in Section \ref{sec:performance_exp}. 
For each model, this results in 5 scores for every metric per dataset. 
Table \ref{tab:expresults_alldatasets} provides the average scores and their standard deviations across the four datasets, with the best performance per metric across models highlighted in bold. 
To ensure that averaging across datasets does not disproportionately reflect the performance of any single dataset, we verified that no individual dataset exhibited an outlier effect on the aggregated results.
For scores specific to each dataset, we refer to the supplementary materials.

\begin{table}[ht]
	\footnotesize
	\centering
	\renewcommand{\arraystretch}{0.8} 
	\caption{\textbf{Average model performance (standard deviation) across all 4 datasets.} Best performance per metric is highlighted in bold.}	
        \begin{tabularx}{\textwidth}{r *{5}{>{\centering\arraybackslash}X}}
			\toprule
			& \textbf{AUC} & \textbf{AP} & \textbf{Brier} & \textbf{relAEC} & \textbf{Savings} \\
			\midrule
			\textbf{boost} & \textbf{0.837 (0.074)} & 0.556 (0.184) & 0.102 (0.050) & 0.288 (0.170) & 0.475 (0.180) \\
			\textbf{csboost} & 0.803 (0.079) & 0.507 (0.175) & 0.202 (0.074) & \textbf{0.554 (0.079)} & \textbf{0.491 (0.121)} \\
			\textbf{logit} & 0.783 (0.027) & 0.475 (0.133) & 0.117 (0.041) & 0.175 (0.083) & 0.343 (0.074) \\
			\textbf{cslogit} & 0.750 (0.034) & 0.371 (0.120) & 0.259 (0.061) & 0.513 (0.051) & 0.431 (0.080) \\
			\textbf{forest} & 0.821 (0.045) & 0.526 (0.131) & 0.115 (0.042) & 0.167 (0.044) & 0.449 (0.091) \\
			\textbf{csforest} & 0.825 (0.054) & 0.531 (0.167) & 0.117 (0.042) & 0.141 (0.036) & 0.438 (0.113) \\
			\textbf{net} & 0.834 (0.063) & \textbf{0.570 (0.193)} & \textbf{0.100 (0.046)} & 0.306 (0.183) & 0.469 (0.169) \\
			\textbf{csnet} & 0.794 (0.034) & 0.433 (0.127) & 0.258 (0.066) & 0.504 (0.046) & 0.425 (0.087) \\
			%\hline
            \midrule
			\textbf{null model} & 0.500 & 0.184 & 0.143 & 0.000 & 0.000 \\
			\bottomrule
		\end{tabularx}
	\label{tab:expresults_alldatasets}
\end{table}

The AUC values, ranging from 0.750 to 0.837, show that all models significantly outperform a random classifier in distinguishing defaulters from non-defaulters. 
However, traditional models like XGBoost, logistic regression, and neural networks outperform their cost-sensitive counterparts.
This is expected, as traditional models focus on minimizing cross-entropy, thus enhancing discriminatory power.
The Average precision (AP) results support this: aside from csforest, cost-sensitive variants underperform compared to the traditional models.
For reference, a random classifier would achieve an AP equal to the fraction of positives, which, averaged across the four datasets, is $(0.0674 + 0.1695 + 0.1995 + 0.3)/4=0.184 $.
Another key difference between traditional and IDCS models is that the latter consistently exhibit poorer performance in terms of the Brier score.
Higher Brier scores for models like \textit{cslogit}, \textit{csboost}, and \textit{csnet} indicate poorer calibration, reflecting larger discrepancies between predicted probabilities and actual outcomes. 
\cite{vanderschueren2022predict} found that while calibration improved the Brier score for cost-sensitive models, it had no effect on other metrics like savings. 
Therefore, calibration was not applied in this study.

In terms of cost metrics, IDCS models consistently outperform standard models, particularly in the relAEC metric.
While the difference in savings is smaller, IDCS models generally exhibit lower variance in savings, reflecting more stable performance.
Notably, although \textit{csforest} performs well in savings, surpassing \textit{cslogit} and \textit{csnet}, it underperforms in relAEC, even compared to cost-insensitive models.
This likely stems from \textit{csforest} being based on \cite{bahnsen2015ensemble}, which has a slightly different approach to cost-optimization by optimizning direct costs in tree splitting rather than AEC.
This highlights the need to evaluate multiple metrics rather than focusing solely on the one being optimized.

\begin{table}[ht]
	\footnotesize
	\centering
	\renewcommand{\arraystretch}{0.8}
	\caption{\textbf{Average rankings across datasets per metric.} Bold indicates best performance. Italics denote significance at the 5\% level, rejecting the null hypothesis of equal performance with the best classifier.}	\begin{tabular}{r>{\centering\arraybackslash}m{2cm}>{\centering\arraybackslash}m{2cm}>{\centering\arraybackslash}m{2cm}>{\centering\arraybackslash}m{2cm}>{\centering\arraybackslash}m{2cm}>{\centering\arraybackslash}m{2cm}}
	\toprule
	 & \textbf{AUC} & \textbf{AP} & \textbf{Brier} & \textbf{relAEC} & \textbf{Savings} & \textbf{Average} \\
	\midrule
	\textbf{boost} & \textbf{2.25} (1.000) & 2.50 (0.665) & 2.25 (0.885) & 4.50 (0.250) & 3.50 & 3.00 ± 0.88 \\
	\textbf{csboost} & 5.50 (0.242) & 5.00 (0.242) & 5.25 (0.303) & \textbf{1.50} (1.000) & \textbf{2.25} & 3.90 ± 1.68 \\
	\textbf{logit} & 6.00 (0.152) & 5.25 (0.173) & 4.25 (0.582) & \textit{6.25 (0.030)} & 7.00 & 5.75 ± 0.94 \\
	\textbf{cslogit} & \textit{7.50 (0.017)} & \textit{8.00 (0.002)} & \textit{7.50 (0.009)} & 2.50 (0.564) & 4.00 & 5.90 ± 2.22 \\
	\textbf{forest} & 3.25 (0.885) & 3.25 (0.665) & 3.00 (0.885) & \textit{6.75 (0.015)} & 3.25 & 3.90 ± 1.43 \\
	\textbf{csforest} & 3.25 (0.885) & 3.75 (0.580) & 4.25 (0.582) & \textit{7.25 (0.006)} & 6.00 & 4.90 ± 1.50 \\
	\textbf{net} & 2.50 (0.885) & \textbf{1.75} (1.000) & \textbf{2.00} (1.000) & 4.25 (0.337) & 4.75 & 3.05 ± 1.22 \\
	\textbf{csnet} & 5.75 (0.173) & \textit{6.50 (0.037)} & \textit{7.50 (0.009)} & 3.00 (0.564) & 5.25 & 5.60 ± 1.50 \\
	\midrule
	$\chi^2_F$ & 17.33 (0.015) & 20.50 (0.005) & 21.50 (0.003) & 20.67 (0.004) & 11.33 (0.125) &  \\
	\bottomrule
	\end{tabular}
	\label{tab:expresults_rankings}
\end{table}

Table \ref{tab:expresults_rankings} shows the average ranking of classifiers across datasets for each evaluation metric.
The rightmost column presents the average performance and standard deviation across metrics for each classifier, while the bottom row provides the Friedman $\chi^2$ statistic, used to assess whether at least two classifiers show significantly different performances.
Additionally, the p-values of a post-hoc test with Hommel p-adjustments that compare each classifier to the best-performing one are added between brackets. 
This method is preferred over Nemenyi's test, which is considered overly conservative and less statistically powerful \citep{garcia2008extension}.
	
The results show that \textit{boost} and \textit{csboost} are the top-performing classifiers on average across metrics. 
Boost leads in AUC, while \textit{csboost} excels in both cost-sensitive metrics.
Notably, the Friedman test indicates significant differences between classifiers for all metrics except Savings, confirming that at least two classifiers perform distinctly. 
Post-hoc analysis with Hommel p-adjustments shows that for AUC, \textit{boost} significantly outperforms only \textit{cslogit}, while for both AP and Brier, \textit{net} significantly outperforms both \textit{cslogit} and \textit{csnet}.
In the case of AEC, \textit{csboost} significantly outperforms \textit{logit}, \textit{forest}, and \textit{csforest}.
Since the Friedman test found no significant differences between classifiers for the Savings metric, no post-hoc analysis was conducted for that metric.

\subsection{Explanation stability}
\subsubsection{Coefficient of Variation}

\begin{figure}[ht]
	\centering
		\begin{subfigure}[b]{0.40\textwidth}
			\includegraphics[width=\textwidth]{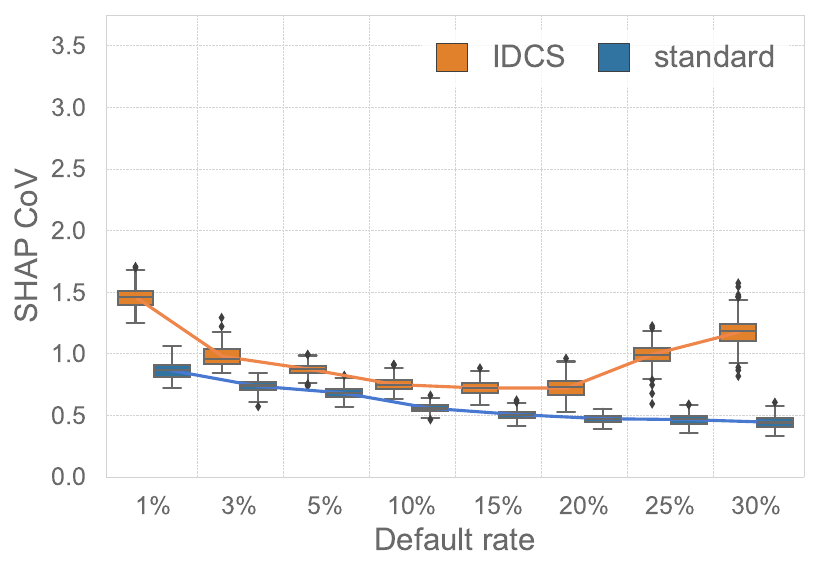}
			\captionsetup{skip=0pt}
			\caption{GMSC}
			\label{fig:GMSC_SHAP_CoV}
		\end{subfigure}
		\begin{subfigure}[b]{0.40\textwidth} 
			\includegraphics[width=\textwidth]{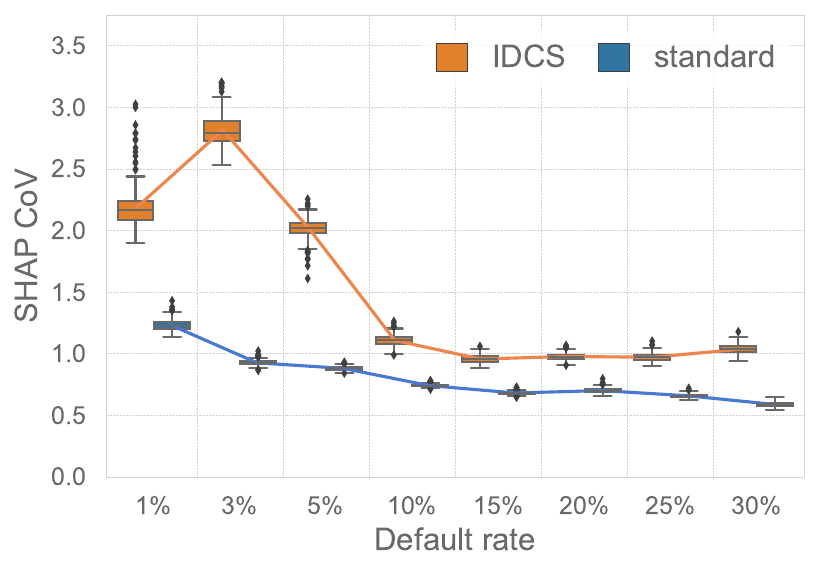}
			\captionsetup{skip=0pt}
			\caption{VUB}
			\label{fig:VUB_SHAP_CoV}
		\end{subfigure}
	\centering
		\begin{subfigure}[b]{0.40\textwidth} 
			\includegraphics[width=\textwidth]{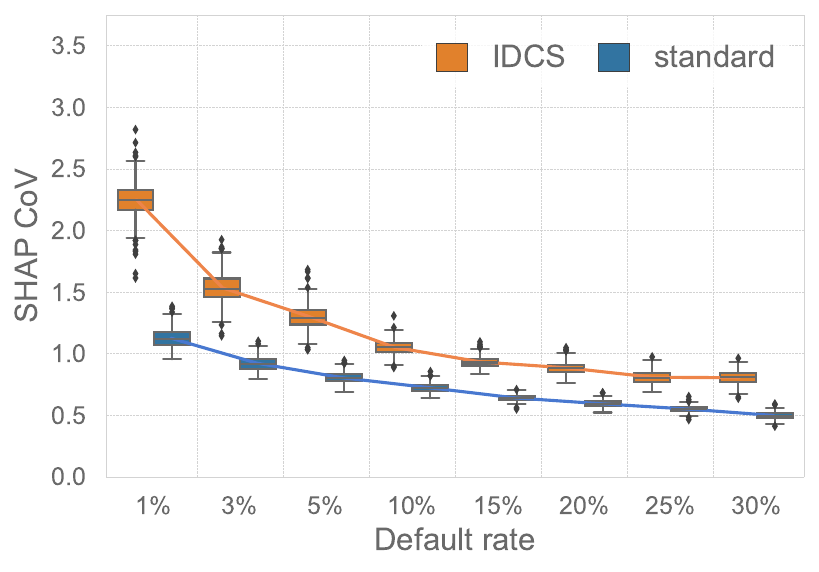}
			\captionsetup{skip=0pt}
			\caption{HMEQ}
			\label{fig:HMEQ_SHAP_CoV}
		\end{subfigure}
		\begin{subfigure}[b]{0.40\textwidth} 
			\includegraphics[width=\textwidth]{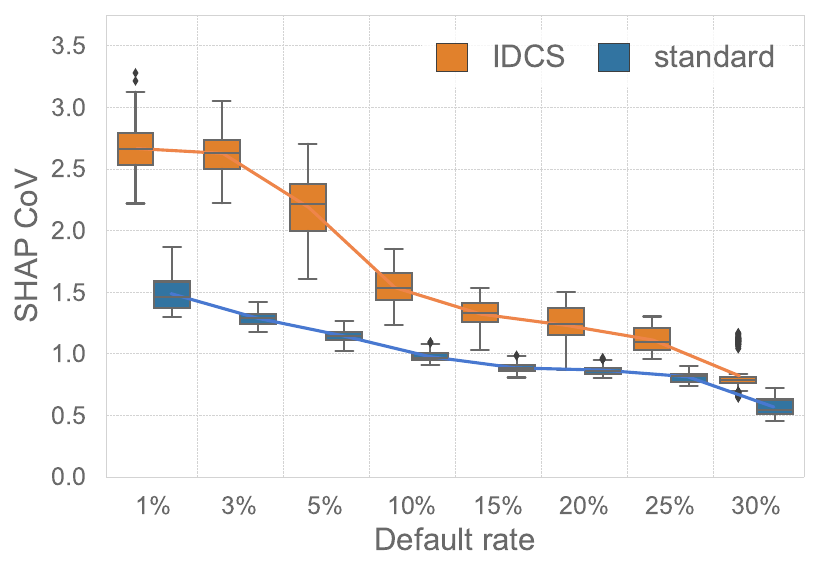}
			\captionsetup{skip=0pt}
			\caption{SGCS}
			\label{fig:SGCS_SHAP_CoV}
		\end{subfigure}
	\captionsetup{aboveskip=5pt, belowskip=-10pt}
	\caption{Distribution of SHAP CoV values}
	\label{fig:SHAP_CoV_stability}
\end{figure}

This section presents the results of the stability experiment, starting with the coefficient of variation (CoV).
The experiment provides SHAP and LIME CoV values for each test observation across all imbalance levels.
To enable a general comparison of IDCS classifiers and traditional models, we average these stability scores for both variants across the four model types.
Figure \ref{fig:SHAP_CoV_stability} illustrates SHAP CoV values for the different datasets.
In these plots, the x-axis represents class imbalance levels (default rates), while the y-axis displays the CoV of SHAP explanations.
Each data point represents the stability of the explanation for an individual test instance, and the boxplot displays the distribution of explanation stabilities of all test instances in the set.
IDCS results are depicted in orange, while traditional results are shown in blue.
For detailed results for each model and dataset, please refer to the supplementary material.

At all imbalance levels, SHAP explanations from IDCS models consistently exhibit higher CoV values than those from cost-insensitive models.
This indicates greater variability in SHAP feature importance values across iterations, resulting in less stable explanations for IDCS classifiers, regardless of the resampled default rate.
The boxplots illustrate a significant lack of overlap at each default rate, emphasizing this difference.
A Two-Sample Kolmogorov-Smirnov (KS) Test confirms that the CoV distribution of IDCS classifiers is significantly higher than that of traditional classifiers at every default rate, with p-values $<$ 0.001 across four datasets (see supplementary materials).

Additionally, the level of class imbalance in the training set influences explanation variability.
Consistent with the findings of \cite{chen2024interpretable} for traditional model variants, SHAP CoV values increase with class imbalance, particularly when the majority-to-minority ratio exceeds 95/5.
Our experiment shows that this effect is even more pronounced for IDCS classifiers.
For every dataset, CoV values rise sharply with increasing class imbalance, indicating greater explanation instability overall and a more significant deterioration in stability under class imbalance.
Notably, for the GMSC dataset, CoV values also increase when rebalancing to a default rate of 25\% or higher, given the original default rate of 6.74\% for this dataset.
\begin{figure}[ht]
	\centering
		\begin{subfigure}[b]{0.40\textwidth}
			\includegraphics[width=\textwidth]{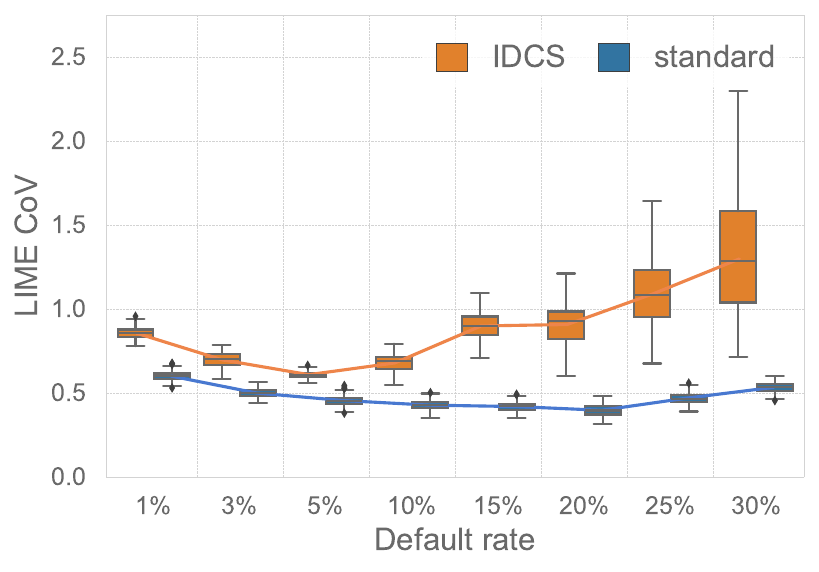}
			\captionsetup{skip=0pt}
			\caption{GMSC}
			\label{fig:GMSC_LIME_CoV}
		\end{subfigure}
		\begin{subfigure}[b]{0.40\textwidth}
			\includegraphics[width=\textwidth]{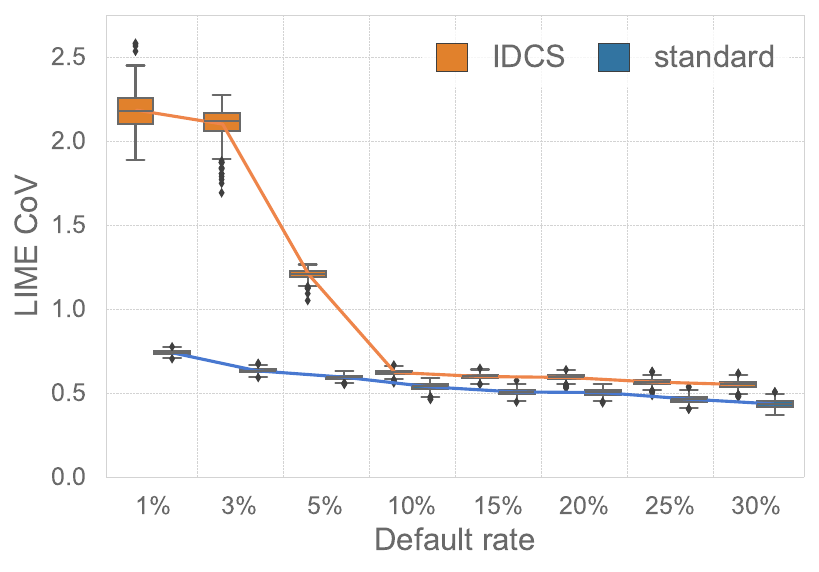}
			\captionsetup{skip=0pt}
			\caption{VUB}
			\label{fig:VUB_LIME_CoV}
		\end{subfigure}
	\centering
	\begin{subfigure}[b]{0.40\textwidth}
			\includegraphics[width=\textwidth]{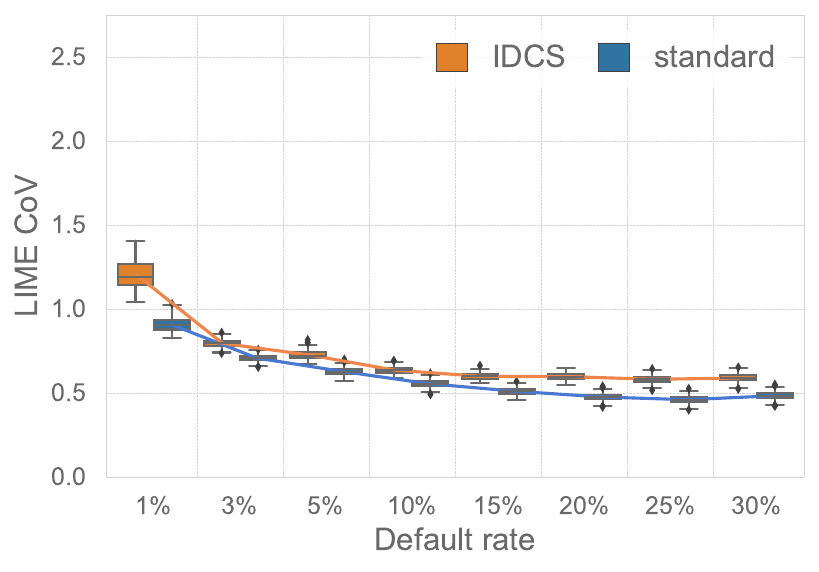}
			\captionsetup{skip=0pt}
			\caption{HMEQ}
			\label{fig:HMEQ_LIME_CoV}
		\end{subfigure}
		\begin{subfigure}[b]{0.40\textwidth}
			\includegraphics[width=\textwidth]{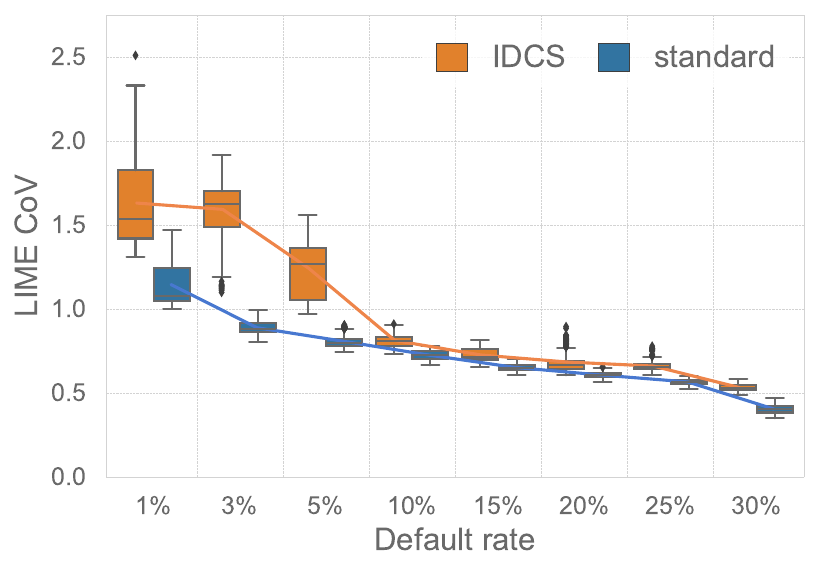}
			\captionsetup{skip=0pt}
			\caption{SGCS}
			\label{fig:SGCS_LIME_CoV}
		\end{subfigure}
	\captionsetup{aboveskip=5pt, belowskip=-10pt}
	\caption{Distribution of LIME CoV values}
	\label{fig:LIME_CoV_stability}
\end{figure}

A similar trend is observed for LIME, as shown in Figure \ref{fig:LIME_CoV_stability}. 
LIME CoV values are consistently higher for IDCS models across all imbalance levels and increase with greater class imbalance.
This indicates that, like SHAP, LIME explanations exhibit increased instability for IDCS classifiers, particularly as class imbalance intensifies. 
The dimensionless nature of the CoV metric enables direct value comparisons across datasets and between SHAP and LIME. 
While some dataset-specific differences exist, LIME CoV scores are generally lower than SHAP CoV scores. 
Mirroring the SHAP results, for the GMSC dataset resampling towards a more balanced dataset also leads to an increase in explanation instability.

\subsubsection{Sequential Rank Agreement}

\begin{figure}[ht]
	\centering
	\begin{subfigure}[b]{0.40\textwidth} % Adjust the width of subfigures
		\includegraphics[width=\textwidth]{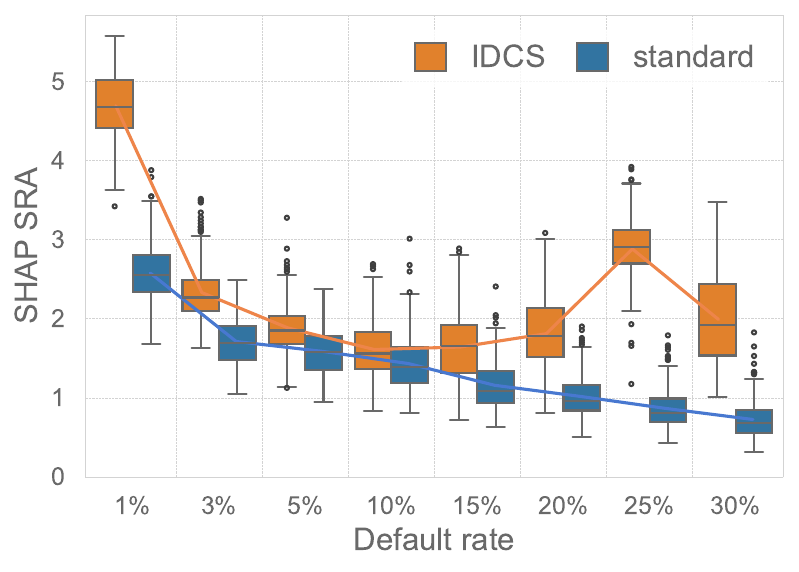}
		\captionsetup{skip=0pt}
		\caption{GMSC}
		\label{fig:GMSC_SHAP_SRA}
	\end{subfigure}
	\begin{subfigure}[b]{0.40\textwidth} % Adjust the width of subfigures
		\centering
		\includegraphics[width=\textwidth]{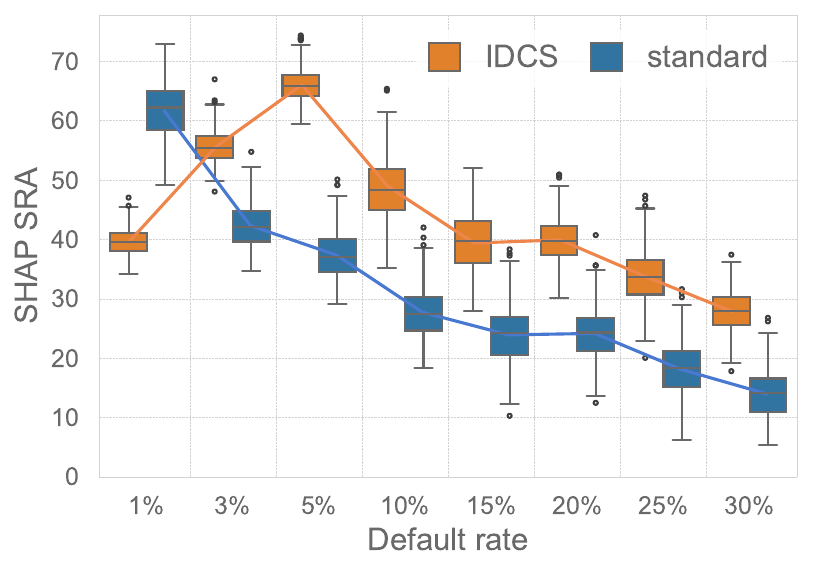} % Use the filename of your second image here
		\captionsetup{skip=0pt}
		\caption{VUB}
		\label{fig:VUB_SHAP_SRA}
	\end{subfigure}
	\centering
	\begin{subfigure}[b]{0.40\textwidth} % Adjust the width of subfigures
		\includegraphics[width=\textwidth]{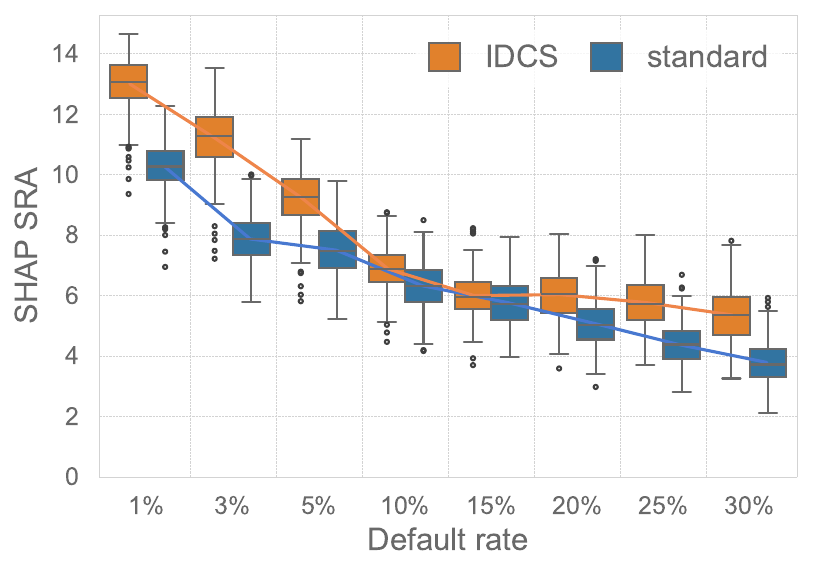}
		\captionsetup{skip=0pt}
		\caption{HMEQ}
		\label{fig:HMEQ_SHAP_SRA}
	\end{subfigure}
	\begin{subfigure}[b]{0.40\textwidth} % Adjust the width of subfigures
		\includegraphics[width=\textwidth]{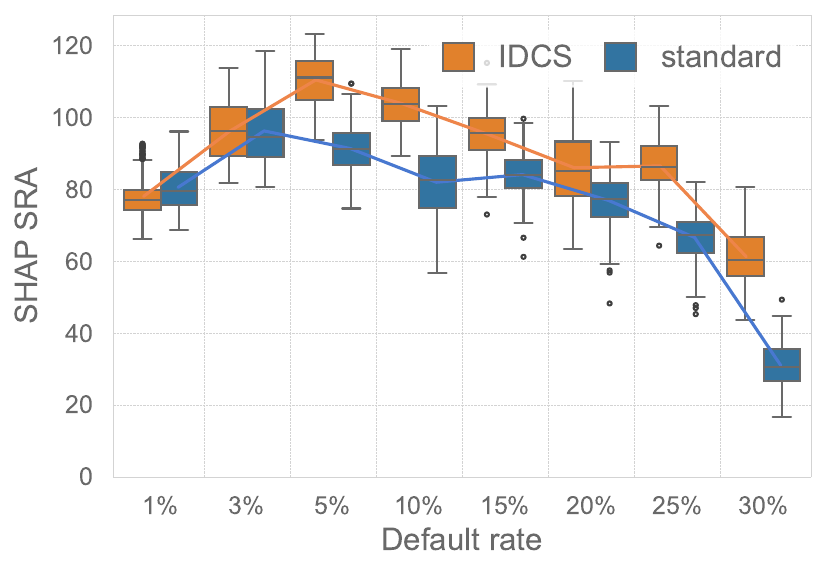} % Use the filename of your second image here
		\captionsetup{skip=0pt}
		\caption{SGCS}
		\label{fig:SGCS_SHAP_SRA}
	\end{subfigure}
	\captionsetup{aboveskip=5pt, belowskip=-10pt}
	\caption{Distribution of SHAP SRA values}
	\label{fig:SHAP_SRA_stability}
\end{figure}

\begin{figure}[ht]
	\centering
	\begin{subfigure}[b]{0.40\textwidth} % Adjust the width of subfigures
			\includegraphics[width=\textwidth]{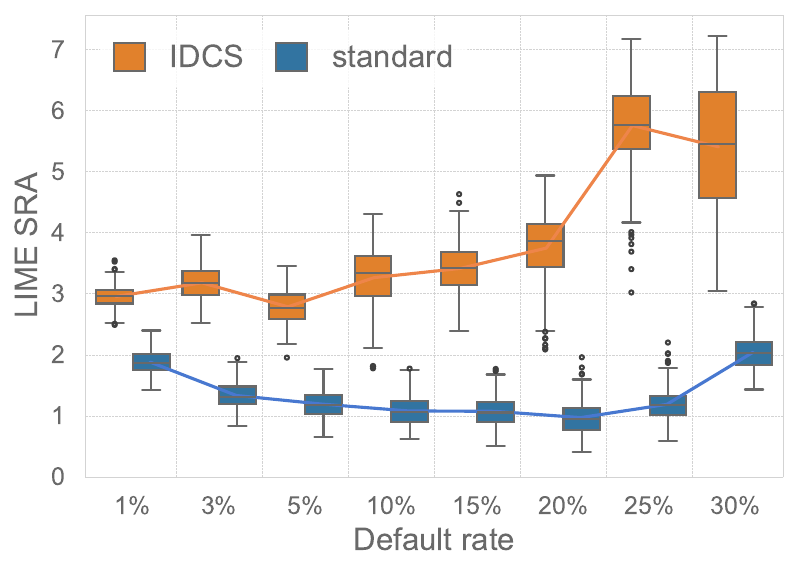}
			\captionsetup{skip=0pt}
			\caption{GMSC}
			\label{fig:GMSC_LIME_SRA}
	\end{subfigure}
	\begin{subfigure}[b]{0.42\textwidth} % Adjust the width of subfigures
		\includegraphics[width=\textwidth]{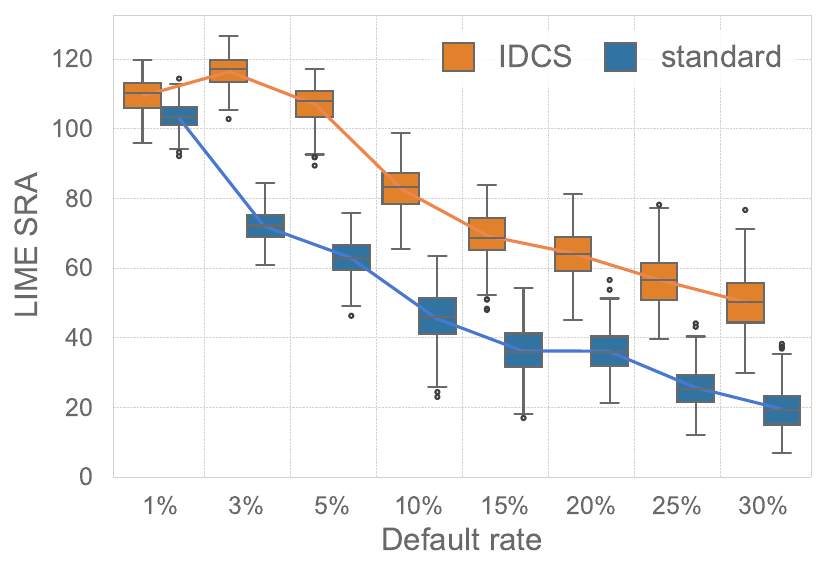} % Use the filename of your second image here
		\captionsetup{skip=0pt}
		\caption{VUB}
		\label{fig:VUB_LIME_SRA}
	\end{subfigure}
	\centering
		\begin{subfigure}[b]{0.40\textwidth} % Adjust the width of subfigures
			\includegraphics[width=\textwidth]{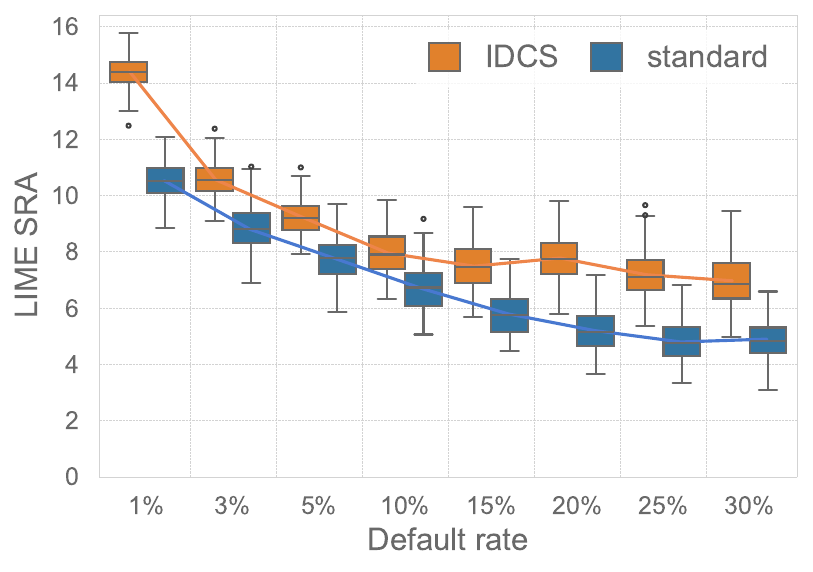}
			\captionsetup{skip=0pt}
			\caption{HMEQ}
			\label{fig:HMEQ_LIME_SRA}
		\end{subfigure}
		\begin{subfigure}[b]{0.40\textwidth} % Adjust the width of subfigures
			\includegraphics[width=\textwidth]{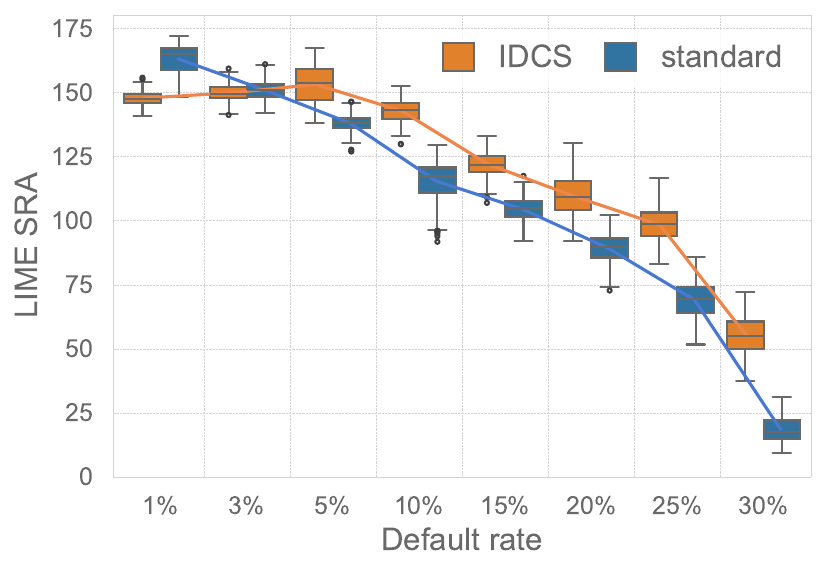} % Use the filename of your second image here
			\captionsetup{skip=0pt}
			\caption{SGCS}
			\label{fig:SGCS_LIME_SRA}
		\end{subfigure}
	\captionsetup{aboveskip=5pt, belowskip=-10pt}
	\caption{Distribution of LIME SRA values}
	\label{fig:LIME_SRA_stability}
\end{figure}

Similar to the CoV, we visualize the Sequential Rank Agreement (SRA) at each imbalance level for SHAP and LIME respectively in Figures \ref{fig:SHAP_SRA_stability} and \ref{fig:LIME_SRA_stability}.
For detailed results on each model and dataset, refer to the supplementary material.
It is important to note that the range of the SRA metric heavily depends on the length of the lists, corresponding to the number of variables in a dataset. 
As a result, the range of the y-axis varies substantially across the four datasets, making direct value comparisons meaningless. 

The findings for SHAP and LIME SRA are consistent with those for SHAP CoV, as IDCS classifiers generally demonstrate significantly higher SRA values across all levels of imbalance, with instability increasing as class imbalance grows, except in a few cases.
For the VUB and SGCS datasets, the effect of class imbalance on SHAP SRA does not show a monotonic increase, and LIME SRA reveals significantly higher instability across all imbalance levels for all datasets except SGCS.

\section{Discussion}\label{sec:discussion}

Our experiments show that IDCS classifiers yield less stable explanations than standard classifiers, especially under high class imbalance. 
We hypothesize that this instability arises from the added complexity of optimizing for instance-dependent costs.
Traditional classifiers estimate the conditional probability $s(\mathbf{x}) = \mathbb{E}(Y \mid \mathbf{x})$ \eqref{eq:condexpvalue}, learning a direct relationship between features $\textbf{x}$ and target $Y$. 
XAI methods like SHAP and LIME rely on this direct relationship, perturbing features to approximate importances.
In contrast, IDCS classifiers optimize $AEC(Y, s(\mathbf{x}), \mathbf{C})$ \eqref{eq:aec}, which introduces a dependency on both feature and cost distributions, as well as their joint distribution.
Consequently, IDCS classifiers entangle features, labels, and costs, as the importance of a feature during training is not only influenced by its presence in the minority class instances but this importance is weighted by the associated costs of those instances, blurring the feature-label relationship.
Perturbing features in SHAP and LIME thus captures not only the influence of $x$ on $Y$ but also the learned interaction with $C$, resulting in less robust explanations.

Our results align with these hypotheses. 
First, we observe higher SHAP and LIME instability with IDCS classifiers at all levels of class imbalance, supporting the idea that incorporating misclassification costs generally makes feature-label relationships noisier.
Second, while the negative effect of class imbalance on SHAP and LIME stability has been observed for traditional classifiers by \citet{chen2024interpretable}, it is more pronounced for IDCS classifiers.
This may be due to the combined effect of a scarcity of positive samples and noisier feature-label relationships, which provide insufficient data for stable post-hoc explanations.
As class imbalance increases, the reduced representativeness of the training sample may amplify explanation instability.

\begin{figure}[ht]
	\centering
	\includegraphics[width=\textwidth]{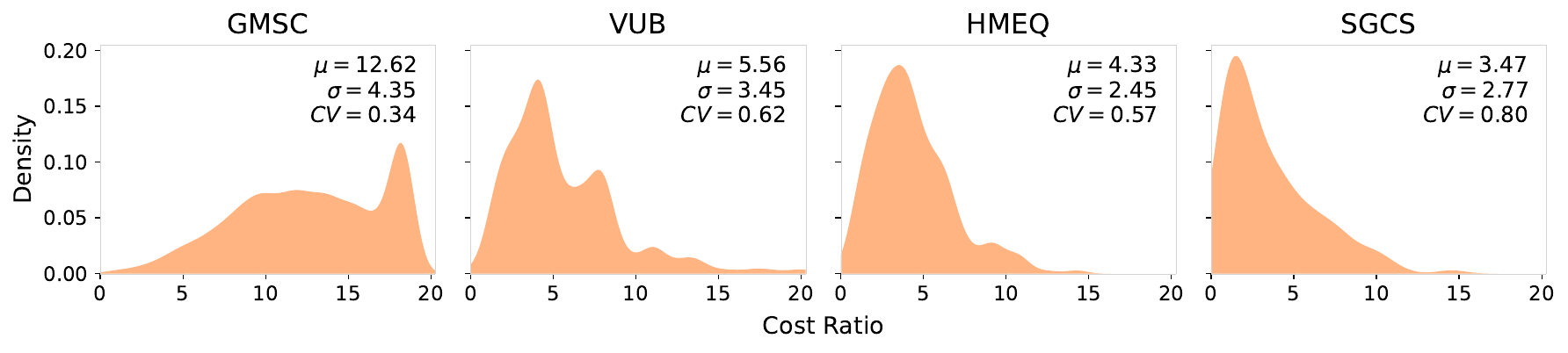}
	\captionsetup{skip=0pt} % Adjust the skip value as needed
	\caption[Cost ratios across datasets]{\textbf{Cost ratios across datasets.}
	The x-axis represents the false negative to false positive $\frac{FN}{FP}$ cost ratio for every actual defualter, derived from Table \ref{tab:cost-matrix-credit}.
	The y-axis shows its normalized frequency, illustrating the distribution of misclassification costs across defaulters.}
	\label{fig:cost_ratios}
\end{figure}

Variations in our results across datasets can be attributed to a combination of factors, including differences in cost distributions, dataset characteristics, and the relationship between costs and features.
Figure \ref{fig:cost_ratios} illustrates the distribution of the false negative to false positive ($\frac{FN}{FP}$) cost ratio, highlighting the greater penalty for granting credit to defaulters compared to the opportunity cost of denial for each instance.
A ratio of 1 signifies equal costs, as assumed in standard classification.
This ratio is central to the AEC metric minimized by IDCS classifiers, and its distribution reflects cost heterogeneity within the dataset.
For example, the GMSC dataset exhibits a relatively smooth distribution, except for a concentration of very high-cost instances.
In contrast, the other three datasets display right-skewed distributions, with high-cost instances acting as cost outliers.
Notably, in the SGCS dataset, around 25\% of instances have a cost ratio close to 1, meaning from a cost-sensitive perspective, they behave similarly to standard classification.
Given that this dataset contains only 1,000 instances, the actual IDCS behavior may be driven by very few cases, exacerbating the discussed issues.
Moreover, the datasets exhibit significant variations in size, the types of variables (categorical vs. numerical), and feature cardinality that may contribute to the differences.

\section{Conclusions and future research}
\label{sec:conclusions}

This paper bridges the gap between two key areas in state-of-the-art credit scoring: model explainability and cost-sensitive learning. 
We conduct a dual-faceted experiment, analyzing both the performance and explanation stability of traditional machine learning models and their instance-dependent cost-sensitive (IDCS) variants. 
Using four open-source datasets, we evaluate model performance in terms of discriminatory power and cost-sensitive metrics, introducing relative AEC (relAEC) as a variant of Average Expected Cost (AEC) that facilitates comparison across datasets.
In line with the theoretical motivation of the models and supporting prior findings \citep{vanderschueren2022predict}, IDCS models outperform on cost-sensitive metrics, while traditional models excel in discriminatory power. 
Notably, \textit{csforest}, the IDCS variant of random forest, achieves competitive AUC and AP, though its relAEC performance is suboptimal due to its different approach to cost minimization. 

For explanation stability, the focus is placed on SHAP and LIME, two commonly used model-agnostic interpretation techniques.
Our analysis (1) compares the stability of explanations between IDCS and traditional classifiers, and (2) examines the effect of class imbalance on this stability. 
Stability is measured using the Coefficient of Variation (CoV) for feature importance and Sequential Rank Agreement (SRA) for feature ranking.
The results show that IDCS classifiers produce significantly less stable model explanations for both SHAP and LIME, regardless of class imbalance.
Additionally, the instability of LIME and SHAP feature importances increases with greater class imbalance, particularly at extreme levels (1\%, 3\%). 
Similar findings were reported by \cite{chen2024interpretable} for cost-insensitive XGBoost and random forest models.
Our results extend these conclusions, showing that this pattern persists across all four model types for both traditional and IDCS classifiers.
Moreover, the negative effect of class imbalance is found to be stronger for IDCS models' explanations, likely because these classifiers depend not only on class distribution but also on the cost distribution of the minority class, making them inherently less stable.

Given the growing adoption of XAI techniques like SHAP and LIME by financial institutions, our research holds significant implications, urging caution in addressing two key challenges in credit scoring: optimizing for costs and ensuring interpretability.
This is particularly important when applying IDCS classifiers to highly imbalanced datasets, which are common in credit scoring. 
Unless the stability issues inherent to IDCS classifiers are addressed, their potential cost efficiency may be compromised by regulatory constraints. 
Therefore, improving explanation stability is essential to ensure both practical application and regulatory compliance of IDCS classifiers.
This is particularly concerning in credit scoring, but also extends to other cost-sensitive areas such as healthcare, customer churn, and fraud detection.

Future work could explore the interplay between cost distributions and feature attributions, using controlled synthetic datasets as done by \citet{de2023robust}, to examine IDCS classifiers' sensitivity to cost outliers.
Additionally, it could investigate whether IDCS learning requires larger datasets or less noisy cost distributions to maintain stability at a given imbalance rate.
Another avenue of exploration is to investigate the specific contribution of the credit scoring cost-matrix used.
On one hand, alternative cost matrices, such as the one proposed by \citet{martin2025novel}, can be explored.
On the other hand, modifying existing cost matrices using instance-based Loss Given Default (LGD) or Exposure at Default (EAD) implementations presents interesting future extensions.
A last logical extension would involve developing new IDCS models or XAI techniques that provide more stable explanations. 
Recent approaches in this direction include S-LIME proposed by \cite{zhou2021s} and RankSHAP introduced by \cite{goldwasser2024provably}, both of which leverage hypothesis testing to enhance the stability of SHAP and LIME explanations.

\section{Acknowledgements}

This research was supported by Flanders Make, the strategic research centre for the Flemish manufacturing industry.

\bibliography{references}

\end{document}